\pdfoutput=1

\documentclass[11pt]{article}

\usepackage{EMNLP2022}

\usepackage{times}
\usepackage{latexsym}
\usepackage{paralist}

\usepackage[T1]{fontenc}

\usepackage[utf8]{inputenc}

\usepackage{microtype}
\usepackage{inconsolata}
\usepackage{amsmath}
\usepackage{amssymb}
\usepackage{bm}
\usepackage{booktabs}
\usepackage{longtable}
\usepackage{multirow}
\usepackage{subcaption}
\usepackage{graphicx}
\usepackage{amsfonts}
\usepackage{soul}
\usepackage{cleveref}

\usepackage{widows-and-orphans}

\newcommand{\ignore}[1]{}
\newcommand{\abr}[1]{\textsc{#1}}

\newcommand{\orange}[1]{\textcolor{OrangeRed}{\textbf{#1}}}
\newcommand{\blue}[1]{\textcolor{RoyalBlue}{\textbf{#1}}}

\newcommand{\lda}[0]{\abr{LDA}}
\newcommand{\mallet}[0]{\abr{Mallet}}
\newcommand{\scholar}[0]{\abr{Scholar}}
\newcommand{\scholarkd}[0]{\abr{Schlr+KD}}
\newcommand{\dvae}[0]{\abr{d-vae}}
\newcommand{\ctm}[0]{\abr{ctm}}

\newcommand{\ah}[1]{{\small\textcolor{orange}{[#1 -- \textsc{hoyle}]}}}
\newcommand{\psrcomment}[1]{{\small\textcolor{olive}{[#1 -- \textsc{psr}]}}}
\newcommand{\pg}[1]{{\small\textcolor{purple}{[#1 -- \textsc{goel}]}}}

\newcommand{\needscite}[1]{\hl{[CITE: #1]}}
\newcommand{\todo}[1]{{\small\textcolor{red}{[\textsc{todo}: #1]}}}
\title{Are Neural Topic Models Broken?}

\author{
    \normalfont
    \noindent\makebox[0pt]{
    \begin{tabular}{@{}c c c c@{}}
        \textbf{Alexander Hoyle} & \textbf{Pranav Goel} & \textbf{Rupak Sarkar} & \textbf{Philip Resnik}\\
        Computer Science & Computer Science & Computer Science & \abr{umiacs}, Linguistics \\
        \multicolumn{4}{c}{University of Maryland} \\
        \multicolumn{4}{c}{\texttt{\{hoyle,pgoel1,rupak,resnik\}@cs.umd.edu}}
    \end{tabular}
    }
}
\begin{document}
\maketitle
\begin{abstract}
\ignore{PSR started putting together ideas here: A line of research dating back to CITE explores the use of neural topic models as an alternative to LDA. Recent work has shown, however, that claims for having advanced the topic modeling state of the art in this literature are not well founded: NPMI, the most widely used evaluation metric, ... 
\emph{Continue reformulating here...}
}

Recently, the relationship between automated and human evaluation of topic models has been called into question. Method developers have staked the efficacy of new topic model variants on automated measures, and their failure to approximate human preferences places these models on uncertain ground. Moreover, existing evaluation paradigms are often divorced from real-world use.

Motivated by content analysis as a dominant real-world use case for topic modeling, we analyze two related aspects of topic models that affect their effectiveness and trustworthiness in practice for that purpose: the \emph{stability} of their estimates and the extent to which the model's discovered categories \emph{align with} human-determined categories in the data. We find that neural topic models fare worse in both respects compared to an established classical method. We take a step toward addressing both issues in tandem by demonstrating that a straightforward ensembling method can reliably outperform the members of the ensemble. 

\ignore{We also offer a simple solution to improve the stability of some newer models.} 
\end{abstract}
\section{Introduction}

\ignore{
When topic models are used in the real world, it is almost always in the service of \textit{automated content analysis}. \todo{go through sample of LDA citations} \todo{Add content analysis definition.}
For resulting insights to be valid, they must have a valid foundation.
Although validity is a multifaceted concept, a model can be considered valid when it (a) is known to produce outputs that align with an existing human standard, and (b) those outputs are are consistent over runs.

\begin{itemize}
    \item No comparison to \mallet{}, a strong baseline
    \item Nobody has looked at these issues comprehensively from the perspective of understanding whether or not models satisfy the needs of the people \emph{doing} content analysis.
    \end{itemize}
  }

Topic models provide an unsupervised way both to discover implicit categories in text corpora, and to estimate the extent to which any given category applies to a specific text item.
As such, topic modeling can be viewed as an automated variety of \emph{content analysis} for text: those two capabilities directly correspond to the practice of developing an emergent \emph{coding system} via examination of a text collection, and then \emph{coding} the text units in the collection \cite{stemler2000overview,smith2000content}.\ignore{As such, topic modeling can be viewed as an automated variety of \emph{content analysis} for text: those two capabilities correspond directly to the widespread practice in content analytic research of empirically developing an emergent \emph{coding system} via examination of a text collection, and then \emph{coding} the text units in the collection \cite{stemler2000overview,smith2000content}.}
This form of content analysis is a dominant use case for topic models and therefore it is our focus here.\ignore{\footnote{There are certainly other use cases for topic models, e.g. extracting latent features for downstream use in text classification. However, a literature review supports this characterization of topic model use cases; see Appendix~\ref{appendix:litsurvey}.}}

Explicitly identifying topic models as a tool for content analysis allows us to characterize what makes topic models \emph{good}: we can measure the extent to which a model achieves the goals of content analysis.\ignore{working from the goals of content analysis to rigorous measurements of the extent to which a model is achieving those goals.}
This careful consideration of the criteria for ``good'' topic models is essential because recent results have challenged the validity of the prevailing model evaluation paradigm \citep{hoyle-etal-2021-automated,doogan-buntine-2021-topic,doogan2022topic,harrando-etal-2021-apples}.
\ignore{call into question the automated evaluations used to support claimed advances in the state of the art using neural methods}
In particular, \citeauthor{hoyle-etal-2021-automated} identified a \emph{validation gap} in the automatic evaluation of topic coherence: metrics like the widely-used normalized pointwise mutual information (NPMI) were never validated using human experimentation for the newer neural models once they emerged, and the authors demonstrated that such metrics exaggerate differences between models relative to human judgments.
Given that the majority of claimed advances in topic modeling are predicated on these metrics (per \citeauthor{hoyle-etal-2021-automated}'s meta-analysis), it would appear that much of the topic model development literature now rests on uncertain ground.
\ignore{\footnote{\psrcomment{I'm not sure the ``fails to track judgments'' wording is exactly right, since the main result involved failing to track model-level preferences, but I'm trying to avoid a deep-dive into what was actually done in that paper; is there better wording that is still self-contained?}}}
\citeauthor{doogan-buntine-2021-topic} also challenged the validity of current automated evaluation measures, and highlighted the disconnect between these measures and use of topic models in real-world settings.\looseness=-1

In this paper, we begin with the needs of content analysis, and we use those needs to argue for specific choices of how to measure topic model performance.
We then report on comprehensive experimentation using two English-language datasets, four neural topic models that are representative of the current state of the art, and classical LDA with Gibbs sampling as implemented in \mallet{} \cite{mallet}.
The results indicate that \mallet{} is a more reliable choice than the more recent neural models from a content analysis perspective.
\ignore{PSR version prior to AH edits... Explicitly identifying topic models as a tool for content analysis allows us to characterize what makes topic models \emph{good}: working from the goals of content analysis to rigorous measurements of the extent to which a model is achieving those goals.  This careful consideration of the criteria for ``good'' topic models is essential because recent results call into question the automatic topic model evaluations that have been used to support claimed advances in the state of the art using neural methods \citep{hoyle-etal-2021-automated}.  In particular, \citeauthor{hoyle-etal-2021-automated} identified a \emph{validation gap} in the automatic evaluation of topic coherence:  normalized pointwise mutual information (NPMI), the most widely used coherence measure, was never validated using human experimentation for neural models, and they demonstrated that exaggerates differences between models relative to human judgments.\ignore{\footnote{\psrcomment{I'm not sure the ``fails to track judgments'' wording is exactly right, since the main result involved failing to track model-level preferences, but I'm trying to avoid a deep-dive into what was actually done in that paper; is there better wording that is still self-contained?}}}
  }
Taking a step toward addressing these issues, we use a straightforward ensemble method that combines the output of models across runs, which reliably yields better results than the usual practice of running a single model.\ignore{\footnote{\psrcomment{Making a note here about the fact that even if you do hyperparameter sweeps, you don't have an objective function to optimize that you can be confident in because of the earlier work invalidating NPMI. That should get discussed somewhere to head off the objection that optimizing hyperparameters is a solution to the instability problem.}}}

To summarize our argument and contributions:
\begin{compactitem}
    \item Automated comparison of topic models should be grounded in a use case, and content analysis is a dominant use case for topic models (\S\ref{subsec:tm_for_content_analysis}).
    \item Stability and reliability are necessary---although not sufficient---criteria to ensure the value of a content analysis (\S\ref{sec:criteria}).
    \item Stability and reliability can be directly measured from model outputs, unlike automated coherence, which prior work has shown is an unreliable proxy for human judgment (\S\ref{sec:metrics}).
    \item On these metrics, we show that LDA with Gibbs sampling (as implemented in Mallet) is significantly more stable and reliable than newer neural models (\S\ref{sec:results}).
    \item We present a straightforward ensembling method to mitigate the stability problem (\S\ref{sec:ensemble}). We release all code and data.\footnote{\url{github.com/ahoho/topics}}
\end{compactitem}

\section{What makes a topic model ``good''?}
\label{sec:background}

In considering how to characterize a topic model that works well, we focus on text content analysis as a dominant use case for topic modeling.

\subsection{Traditional content analysis}
\label{sec:traditional}

Although content analysis is an extremely broad concept \cite{krippendorff2018content}, a very widely used paradigm across many disciplines is a manual process of inductive discovery of codesets via \emph{emergent coding} \citep{stemler2000overview}, which ``allows categories to emerge from the material without the influence of preconceptions'' \citep{smith2000content}.  \citet{weber1990basic} describes a ``data-reduction process by which the many words of texts are classified into much fewer content categories,'' and this invocation of data reduction in a manual setting provides a sense of why topic modeling, a dimensionality-reduction technique, can be a good fit when considering ways to automate the process.\footnote{This process of inductive category discovery contrasts with the use of pre-existing categories, e.g. those coming from relevant theory, and from the use of ``manifest'' or directly observable characteristics of text. Discussions in the literature often distinguish ``quantitative'' from ``qualitative'' content analysis, with the inductive process we describe being associated with the latter category. This terminological distinction may be overly sharp, however; see \citet{schreier2012qualitative} for useful discussion of relationships and differences among quantitative content analysis, qualitative content analysis, and other forms of qualitative research.}

Typically the inductive process involves multiple researchers independently reading samples of the text units being analyzed, and proposing categories (usually called ``codes'') that they see as present and relevant; they then reconcile their independent proposed categories to produce a candidate codeset with associated definitions and coding guidelines. The candidate codeset is then used by two or more people to independently code (i.e. manually label) a sample of the data, and inter-coder reliability is measured using a chance-corrected agreement measure like Krippendorff's $\alpha$ \citep[cf.][]{artstein-poesio-2008-survey}. If an acceptable level of reliability has not yet been achieved, the codeset and coding guidelines are revisited and revised, and another iteration of independent coding and reliability measurement takes place.  Once reasonable reliability has been achieved, the final set of categories is considered to reflect true structure underlying the text collection. Sometimes the texts in the collection are then manually coded using those categories in order to support quantitative analysis --- possibly with further inter-coder reliability measurement for quality control --- although sometimes the set of categories itself is the intended result, not item-level coding.

\subsection{Topic modeling for content analysis}\label{subsec:tm_for_content_analysis}

\begin{table*}[ht]
    \resizebox{\textwidth}{!}{
    \begin{tabular}{lll}
    \toprule
    \textbf{Model Type} & \textbf{Run} & \textbf{Top Words} \\
    \midrule
    \mallet{} & base & \orange{storm} \orange{tropical} \orange{hurricane} \orange{winds} \orange{depression} \orange{mph} \orange{september} \orange{damage} \orange{cyclone} \orange{system} \\ \ignore{ august day season rainfall later near caused october wind moved low landfall west reported pressure \\}
     & nearest & \orange{storm} \orange{tropical} \orange{hurricane} \orange{winds} \orange{depression} \orange{mph} \orange{september} \orange{damage} \orange{cyclone} \orange{system} \\ \ignore{ august day season rainfall later near october caused moved wind low landfall west reported area \\}
     & median & \orange{tropical} \orange{storm} \orange{hurricane} \orange{depression} \orange{winds} \orange{september} \orange{cyclone} \orange{mph} \orange{system} august \\ \ignore{ season damage day later rainfall october moved low near west wind caused pressure north weakened \\}
     & farthest & \orange{tropical} \orange{storm} \orange{depression} \orange{hurricane} \orange{cyclone} \orange{system} \orange{season} \orange{winds} \orange{september} \orange{mph} \\ \ignore{ day low august west wind october later moved convection developed intensity storms south formed pressure \\}
     \hline
    \dvae{} & base & \blue{tropical} \blue{mph} \blue{storm} \blue{hurricane} \blue{winds} \blue{cyclone} \blue{extratropical} \blue{utc} \blue{rainfall} \\ \ignore{ depression dissipated northwestward season inhg flooding mbar convection weakened shear northeastward northeast damage trough dissipating \\}
     & nearest & \blue{tropical} \blue{cyclone} \blue{hurricane} \blue{storm} \blue{winds} \blue{landfall} depression dissipated convection \blue{extratropical} \\ \ignore{ utc rainfall shear weakened storms trough mph intensified intensity cyclones flooding inhg weakening northwestward circulation \\}
     & median & convection \blue{landfall} shear nhc utc \blue{tropical} mbar northwestward \blue{cyclone} \blue{extratropical} \\ \ignore{ inhg dissipated rainfall hurricane intensification national\_hurricane\_center gusts outages cyclones advisories winds flooding saffir storm northeastward \\}
     & farthest & dvorak southwestward depressions dissipation intensifying conventionally southeastward \\ \ignore{ intensify dissipating anticyclone  weakening dissipated monsoon intensification degenerated shear weaken typhoon trough tropical disturbance convection cyclones advisories gusty \\}
    \bottomrule
    \end{tabular}
    }
    \caption{
      Sets of \textsc{weather} topics for two model types for different runs with different hyperparameters on a Wikipedia dataset, represented in conventional fashion using the most probable ten words per topic. The table visually illustrates \mallet{}'s dramatically greater stability: the top words from the base topic appear in corresponding topics across the the full range of the other nine runs (overlap with base topic in \orange{orange}), while for \dvae{}, a neural topic model, consistency with the base topic begins to show a significant drop-off even with the nearest topic (overlap in \blue{blue}). See \S\ref{sec:results} and \S\ref{sec:closereading} for discussion. 
    }\label{tab:top-word-comparison}
\end{table*}

\ignore{
  The models of interest in this paper are exemplified by Latent Dirichlet Allocation \cite[LDA,][]{blei2003latent}, within which each document $d$ is represented as an admixture of $K$ latent global distributions over words $\bm{\beta}_k \in \Delta^{|V|-1}$, each weighted by a local latent distribution over those $K$ topics $\bm{\theta}_d \in \Delta^{K-1}$.
}
The models of interest in this paper are exemplified by Latent Dirichlet Allocation \cite[LDA,][]{blei2003latent}, within which each of $N$ documents $d$ is represented as an admixture $\bm{\theta}_d$ of $K$ topics, and each topic is itself represented as a distribution $\bm{\beta}_k$ over the vocabulary $V$. Topics can thus be viewed in two complementary ways, as ranking either the words in the vocabulary or the documents in the collection.  These views can be interpreted as corresponding closely to two central elements of a traditional text content analysis. First, the rows in the topic-word distributions matrix $\textbf{B} \in \mathbb{R}^{K\times|V|}$ constitute an inductively determined set of categories analogous to a human-determined codeset; for example, the presence of a topic with top (most probable) words \texttt{artist, museum, exhibition} might correspond to a human analyst identifying the code {\sc art}.  Second, the columns of the document-topic matrix $\bm{\Theta} \in \mathbb{R}^{N\times K}$ constitute a soft coding of documents using the categories in $\textbf{{B}}$.\footnote{This could be converted into traditional discrete coding in a number of ways, e.g. assigning the most probable topic for a document as its code.}
To help illustrate the first step, \ignore{--- and also preview the importance of model stability --- }Table~\ref{tab:top-word-comparison} shows the top words from inferred topic-word distributions $\bm{\beta}_k$ for two model types over multiple runs. 

\paragraph{Reviewing the use of topic models.} Bearing these correspondences in mind, we reviewed the literature to confirm our subjective impression that text content analysis is indeed the dominant use case for topic modeling.\footnote{This review is in the spirit of \citet{liberati2009prisma}, although we are not striving for that level of formality. See Appendix~\ref{sec:app:lda-meta} for more details.}  Using Semantic~Scholar (\url{semanticscholar.org}), we collected research studies \emph{outside} the field of computer science published in 2019--2022 that cite \citet{blei2003latent}, and selected 50 at random. We excluded studies that cite \citeauthor{blei2003latent} but do not actually use any topic model, as well as studies that do not involve language data. We retain those that employ topic model variants, such as STM \citep{roberts2013structural}.
Using Semantic Scholar's reported field of study, disciplines represented include medicine, sociology, business, political science, psychology, economics, and history. We find that 94\% of the papers use a topic model for inductive discovery of categories for human consumption, 68\% of which go on to actually assign human-readable code labels to topics; and 64\% of papers use document-topic probabilities as a form of coding for individual text units. We interpret these results as strongly suggesting that, outside topic model development, the primary \emph{use} of topic models is an automated form of text content analysis as characterized in \S\ref{sec:traditional}.

\subsection{Criteria for good content analysis}\label{sec:criteria}

Having established text content analysis as a central topic modeling use case, we consider criteria for ``good'' analysis motivated by that use case. These then inform the selection of topic model evaluation metrics in \S\ref{sec:metrics}, helping to ensure a correspondence between the way topic models are evaluated and the reasons people are using them.

One key issue in content analysis is \emph{stability} or \emph{intra-coder reliability}:  if the same coder were to look at the same data again (say, separated by a long interval to achieve some degree of independence), would they produce the same results?  When an individual coder cannot produce stable output,\ignore{looking at data multiple times,} this calls into question the quality of the results they have produced any one of those times. 

A second central concern in content analysis is \emph{reproducibility} or \emph{inter-coder reliability}: do two or more independent coders looking at the same data agree with each other?  In the absence of externally provided coding to compare against, what establishes trust in categories or coding is consensus, what \citet{weber1990basic} refers to as ``the consistency of shared understandings'' between coders.

A third concern that is often discussed is \emph{validity}: do categories or measurements actually correspond to whatever they are intended to measure \citep{rubio2005content}?  As \citet{weber1990basic} notes, in content analysis this often goes only as far as face validity, i.e. a subjective perception that a measure (or category) appears to be valid. In contrast, \citet{shapiro1997matter} argue that content analysis ``is only valid and meaningful to the extent that the results are related to other measures''.

Research in content analysis typically focuses on these three issues --- stability, reproducibility, and validity --- as necessary considerations when considering whether a content analysis should be used as the basis for inferences about a dataset.  Validity, however, is challenging to assess outside the context of specific research questions \citep[see][for an example in political science]{grimmer2013text}. We therefore focus on stability and reproducibility as the basis for developing metrics to assess topic models for the automated content analysis use case. 

Note that the criteria we emphasize---stability and reproducibility---are necessary to ensure the value of a content analysis, but not sufficient: topic coherence is a complementary and crucial concern \cite{newman-etal-2010-automatic} that requires further investigation, since prior work has shown automated coherence measurements are an unreliable proxy for human judgment \cite{hoyle-etal-2021-automated}. 

\ignore{
In applying them to topic models, we conceptualize a model as playing the same role as a human content analyst or coder.
}

\subsection{Operationalizing the criteria}\label{sec:metrics}

Because they are generative models, the development community initially evaluated topic models using held-out perplexity, i.e. their ability to predict unseen text. However, focusing on the goal of producing categories that humans can understand, \citet{changReadingTeaLeaves2009} established that perplexity actually correlated \emph{negatively} with human determinations of coherence as estimated using behavioral measures.  \citet{lau-etal-2014-machine} went on to introduce NPMI as an automated coherence metric positively correlated with human preferences. Since then, NPMI has been the most prevalent way to establish that a new topic modeling method is better than the old ones, including the new generation of neural topic models. However, \citet{hoyle-etal-2021-automated} recently identified a \emph{validity gap} for NPMI: its correspondence to human judgments was never validated for neural topic models, and although recent neural topic models can attain relatively high NPMI, human annotators fail to meaningfully distinguish them from a classical LDA baseline.

That result suggests taking a fresh, well motivated look at topic model evaluation. Any model evaluation should be grounded in consideration of the model's intended purpose, which leads us to suggest grounding formal evaluation metrics in the content analysis use case.\footnote{It should be noted that we are focusing on only the most central part of the content analysis use case. \citet{smith2000content} situates codeset discovery and coding within a broader process that begins with identifying the research problem, selecting appropriate materials, etc., and ends with actually using the codeset and coding to generate research findings. \citet{ldawild} situate topic model creation within a corresponding end-to-end workflow; see also \citet{boyd2014care} for practical discussion of topic modeling including discussion of other use cases.}

\subsubsection{Stability}\label{sec:background:stability}

\S\ref{sec:criteria} notes \emph{stability} as an important criterion in content analysis.  Whether codes are being produced by a human coder or a topic model, if there is meaningful latent structure in the text collection, one would expect either humans or models to consistently uncover that structure.\ignore{---that is, to produce the same result if going through the same process multiple times.\ah{Hyperparameter settings do mean that it is not quite the ``same'' process; }}
\ignore{\ah{REDO THIS SENTENCE} Since any particular topic model $\langle\bm{\textbf{B}},\bm{\Theta}\rangle$ exists within a large space of \emph{possible} models, depending on a range of hyperparameter settings and possibly randomized parameter estimation methods and manual human evaluations is intractable in such a large space, we focus on selection of \emph{automated} measurements.}

To ground our evaluation in our use case, we measure the stability of models across hyperparameter settings (for a fixed topic number $K$). 
In the absence of an unsupervised metric to optimize or reliable ``default'' values, a practitioner is forced to explore different hyperparameter settings.
All else equal, a topic model that is less sensitive to changes in hyperparameter settings is preferable to one that is more sensitive (we also evaluate results for fixed hyperparameters with different random seeds, see Appendix~\ref{app:fixed-hparam}).

Translating these ideas into a formal measurement, we follow \citet{greene2014many} in operationalizing model stability by measuring the total distances between the topic-word estimates for each run, extending their method to measure stability of both the sorted rows of the topic-word estimates ${\textbf{B}}$ or the sorted columns of the document-topic estimates $\bm{\Theta}$; the smaller these distances, the more stable the estimates.

Without loss of generality, we focus on the topic-word distributions to operationalize stability as \emph{total topic distance}.
We collect a set of $\bm{\beta}^{(i)}_{k}, i\in 1\ldots,m, k\in 1\ldots K$ estimates from $m$ model runs on the same dataset.
For each pair of $m \choose 2$ runs, we compute the pairwise distance $d$ between all $K$ topics in each run. 
We use the Rank-Biased Overlap distance \citep[RBO,][]{webber2010similarity}, which is used to measure the distance between two rankings giving more importance to similarity of the top-ranked items, i.e., the measure is \emph{top-weighted}, making it ideal for measuring the distance between topics \citep{mantyla2018measuring}.\footnote{Experiments with distances that used the full distribution, like Jensen-Shannon divergence, led to matched topics that were less interpretable.}
Within a pair of runs $\textbf{B}^{(i)}, \textbf{B}^{(j)}$, the goal is to find a permutation of rows $\pi(\cdot)$ to minimize
\begin{align}
    \mathcal{TD}(\mathbf{B}^{(i)},\mathbf{B}^{(j)}) = \frac{1}{K}\sum_k d\left(\bm{\beta}_k^{(i)}, \bm{\beta}_{\pi(k)}^{(j)}\right)
\end{align}
This problem is an instance of bipartite matching distance minimization, which we solve with the modified Jonker-Volgenant algorithm of \citet{crouse2016}.
If the set of $m \choose 2$ total distances $\mathcal{TD}$ (i.e., the minimized costs) for one model are significantly less than a second model, the first model is more stable. 

Prior topic modeling work has identified stability as a crucial concern for robust application to social sciences \citep{koltcov2014latent,ballester2022robustness}, for better incorporation of topic models in downstream automated NLP tasks \citep{miller-mccoy-2017-topic}, as a criterion for tuning LDA parameters \citep{greene2014many}, and has offered ways to improve it for LDA estimates \citep{agrawal2018wrong,mantyla2018measuring}.
\citet{chuang-etal-2015-topiccheck} introduced an interactive tool to help humans assess a topic model's stability. However, in a meta-analysis of 35 papers proposing new ``state-of-the-art'' neural topic models over the past three years (2019-2022), we find that \emph{none} of them compared the models on stability.\footnote{We select publications from the meta-analysis in \citet{hoyle-etal-2021-automated}, updated with recent work from the most common venues in that list. Although papers do not measure the stability of the estimates directly as we do here, six papers do report variance over chosen metrics.}

\subsubsection{Inter-coder reliability}\label{sec:background:cluster}

\S\ref{sec:criteria} notes that \emph{reproducibility} or \emph{inter-coder reliability} is also a central  consideration in content analysis. Going beyond intra-coder consistency, if a set of codes cannot be applied consistently by multiple coders, this also calls into question whether it is doing a good job capturing meaningful content categories.

We treat a topic model $\langle{\textbf{B}},\bm{\Theta}\rangle$ as a coder, and approach inter-coder reliability from the perspective of reproducing categories from other coders who are human, instantiated as a set of human-assigned ``ground truth'' labels for the documents in the collection. Since what we care about here are the categories discovered by a topic model, not actual labels, we measure the extent to which categories induced by the model \emph{align with} that ground truth.  Intuitively, for example, if documents that are assigned to a topic by the model all have the same ground-truth label, the topic is a good fit for human categorization of the data (and this can be determined just using documents assigned to the topic, without any generation or evaluation of labels). Conversely, if documents all assigned to the same topic in the model have a wide variety of ground truth labels, this mismatch suggests that the topic is missing something important relative to the underlying category structure in the collection.

By taking the most probable topic for a document $\hat{\ell_d} = \text{argmax}_{k'}\ \theta_{d,k'}$ as its assigned topic or ``code'', we can apply standard metrics of cluster quality\ignore{\needscite{TM paper that first proposed this. Nguyen 2015?}}.\footnote{It is common in the topic model development literature to evaluate models by learning a mapping $f: \bm{\theta}_d \to \ell_d$ from training data and calculating a held-out $F_1$ score---i.e., to train a classifier---but this process does not correspond to any common real-world use of topic models.} We borrow exposition of cluster quality metrics from \citet{poursabzi-sangdeh-etal-2016-alto}, with all metrics using the predicted clustering from a model, $\hat{\mathcal{L}} = \{ \ell_d : d=1,\ldots,n \}$, and a given set of gold labels $\mathcal{L}^*$.

\paragraph{Adjusted Rand Index.} The Rand Index compares all pairs of the two labelings over documents, counting the proportion of pairs that have the same (TP) or different (TN) assignments \cite{rand1971index}.
\begin{align*}
    \text{RI} = \frac{TP + TN}{TP + FP + TN + FN} 
\end{align*}
The adjusted rand index further corrects for chance \cite{steinley2004adjustedRand}.

\paragraph{Normalized Mutual Information (NMI)} measures the mutual information between two clusterings, and is invariant to cluster permutations \cite{strehl2003nmi}. Here, $\mathbb{I}$ is the mutual information and $\mathbb{H}$ are the entropies for each clustering.
\begin{align}
    \text{NMI} = \frac{2\mathbb{I}\left(\hat{\mathcal{L}}, \mathcal{L}^*\right)}{\hat{\mathbb{H}} + \mathbb{H}^*}
\end{align}

\paragraph{Purity} takes all documents contained in a single \emph{predicted} cluster and measures the number of associated gold labels that appear in it --- it is roughly akin to precision \cite{zhao2001purity}. 
A small number of gold labels present in a predicted cluster means that there is high alignment between the discovered ``concept'' and the true one. 
\begin{align}
    P(\hat{\mathcal{L}}, \mathcal{L}^*) = \frac{1}{n} \sum_k \text{max}_{k'} | \hat{\mathcal{L}}_k \cap {\mathcal{L}}^*_{k'} |
\end{align}
With $\mathcal{L}_k = \{ \ell_d : d=1,\ldots,n; \ell_d = k \}$.
Purity is not symmetrical, so we define \emph{inverse purity} as $P^{-1} = P(\mathcal{L}^*, \hat{\mathcal{L}})$, and $P_1$ as their harmonic mean (analogous to $F_1$).

Prior topic modeling work has looked at how well topics discovered by a model align with reference codes \citep{chuang2013topic,korenvcic2021topic}. However, in the same meta-analysis discussed above, only \emph{six} of the 35 neural topic modeling development papers compared models on a version of alignment. This suggests that even though stability and alignment have been identified as important and useful criteria in topic modeling literature in prior work --- especially when using and examining LDA and its variants --- they have seen precious little uptake. We hope that our strong use-case motivations and experimental results will change this. 
\section{Experiments}
\label{sec:experiments}

Having argued that topic models should be subject to evaluations designed with real-world uses in mind, and having motivated specific ways to operationalize evaluative measurements based on criteria that matter for text content analysis, we evaluate nominally ``state-of-the-art'' topic models to understand how well they perform relative to those criteria.

\subsection{Datasets}
We use two standard English datasets of varying characteristics: 14,000 ``good'' articles from Wikipedia \cite[\texttt{Wiki},][]{merityPointerSentinelMixture2017} and 32,000 bill summaries from the 110-114th U.S. congresses (\texttt{Bills}).\footnote{``Featured'' Wikipedia articles have an incompatible labeling scheme and are therefore excluded.  Raw bill data was extracted from \url{https://www.govtrack.us/data/us/}.} 
 The documents in both datasets have hierarchical labels, which serve as ground truth when evaluating the quality of the document-topic posteriors (\S\ref{sec:metrics}).
The \texttt{Wiki} dataset has 45 labeled high-level and 279 low-level labels; the \texttt{Bills} dataset has 21 high-level and 114 low-level labels.
We process each with the standardized setup of \citet{hoyle-etal-2021-automated}, setting the vocabulary size to either 5,000 or 15,000 terms, limiting by term-frequency \citep{blei2006correlated}.

Prior evidence suggests that neural topic models may produce topics with narrower scope than classical models \citep[e.g., \texttt{agnes\_martin, sol\_lewitt, minimalism} rather than \texttt{art, painting, museum}, cf.][]{hoyle-etal-2021-automated}.
We therefore generate held-out sets for both datasets to facilitate exploration of this phenomenon.
Namely, we ensure that both the training and held-out sets contain documents from all \emph{high-level} categories, but partition the \emph{low-level} categories into seen and unseen labels.
For example, Wikipedia articles about \texttt{television} are present in both subsets, but those about \texttt{30 Rock} episodes are exclusively in the training set whereas \texttt{Simpsons} episodes are unseen.
Although not an emphasis of the present work, our high-level conclusions remain the same for the held-out data (i.e., \mallet{} is better-aligned, Appendix~\ref{app:test-alignment}); we leave further analysis to future efforts.

\ignore{
Prior evidence suggests that neural topic models may produce topics with narrower scope than classical models \citep[e.g., \texttt{martin, lewitt, minimalism} rather than \texttt{art, painting, sculpture},][]{hoyle-etal-2021-automated}.
To better measure this effect, we introduce a novel strategy to generate a hold-out set \todo{check novelty}.
Namely, we ensure that both the training and hold-out sets contain documents from all \emph{high-level} categories, but partition the \emph{low-level} categories into seen and unseen labels.
That is, Wikipedia articles about \texttt{television} are present in both subsets, but those about \texttt{30 Rock} episodes are exclusively in the training set whereas \texttt{Simpsons} episodes are unseen.
We expect that models that have generalizable topic estimates will align better with ground truth on the unseen data.
}

\subsection{Models and experimental contexts}
\emph{Classical} topic models use Gibbs sampling or variational inference to infer the posteriors over the latent variables; more recent \emph{neural} topic models use contemporary techniques that involve neural networks, such as variational auto-encoders \citep{Kingma2014AutoEncodingVB}.

We evaluate one classical topic model and four neural topic models. Each model is evaluated in one of 16 \emph{experimental contexts}: a tuple of dataset (\texttt{Bills}, \texttt{Wiki}), vocabulary size (5k, 15k), and number of topics (25, 50, 100, 200).

In light of the finding that automated coherence cannot meaningfully reproduce human judgments \cite{hoyle-etal-2021-automated}, there is no unsupervised metric that we can optimize to avoid the problem of instability, while optimizing for $K$ remains an open research problem.
Therefore we vary $K$ and, for all contexts, we train the models ten times using a different set of randomly-selected hyperparameters, where value ranges are based on prior literature (\S\ref{sec:app:hparams}).

Although ``optimal'' hyperparameters will often change depending on context, we also report results with fixed hyperparameters and varying seeds in Appendix~\ref{app:fixed-hparam}.

\paragraph{\mallet{}.} Given its prevalence among practitioners and strong qualitative human ratings in prior work \cite{hoyle-etal-2021-automated}, as a classical model we use LDA estimated with Gibbs sampling \cite{griffiths2002Gibbs}, implemented in \mallet{} \cite{mallet}. While LDA is a common baseline in the topic model development literature, it is often estimated with variational methods, which anecdotally produce lower-quality topics \cite{goldberg-tweet}.\footnote{\citet{mimno-spacy} provides discussion of why stochastic variational Bayes, which has seen widespread use in topic modeling using the {\tt gensim} library (\url{https://radimrehurek.com/gensim/}), may be particularly problematic.}

\paragraph{\scholar{}.} A popular neural alternative to the structural topic model \cite{roberts2014structural}, flexibly incorporating supervised signals and external covariates into the model \cite{card-etal-2018-neural}.

\paragraph{\textsc{Scholar+KD}.} \ignore{Many neural models are adaptable: it is not necessary to re-derive an inference procedure when the generative model changes, allowing for \hl{novel model variants} that would not be tractable in classical settings. \cite{hoyle-etal-2020-improving}} \citet{hoyle-etal-2020-improving} apply knowledge distillation (KD) to improve on \abr{\scholar{}} using a \abr{bert}-based autoencoder. \citet{gao-etal-2021-topic} show that domain experts prefer the outputs of an adapted \textsc{Scholar+KD} over other models \cite[\mallet{}, \abr{ETM},][]{dieng2020topic}.

\paragraph{Dirichlet-VAE.} The Dirichlet-VAE \cite[D-VAE,][]{burkhardtDecouplingSparsitySmoothness2019} is a variant of LDA that (a) uses a \abr{vae} to approximate the posterior over the latent document-topic distribution, and (b) follows \abr{ProdLDA} by using unnormalized estimates of the topic-word values $\beta$, as opposed to a proper distribution. Annotators rate \abr{d-vae}'s topics similarly to \mallet{} \cite{hoyle-etal-2021-automated}.

\paragraph{Contextualized Topic Model.} Typically, \abr{vae}-based neural topic models encode the bag-of-words representation of a document with a neural network to parameterize that document's distribution over topics. The popular model introduced in \citet{bianchi-etal-2021-pre} extends this representation with a contextualized document embedding from a large pretrained language model.

\section{Results}
\label{sec:results}

\begin{table*}[ht]
    \resizebox{\textwidth}{!}{
    \begin{tabular}{llrrrrrrrrrrrrrrrr}
    \toprule
     &  & \multicolumn{8}{c}{$|V| = $ 5k} & \multicolumn{8}{c}{$|V| = $ 15k} \\
    \cmidrule(lr){3-10}\cmidrule(lr){11-18}
     &  & \multicolumn{2}{c}{$k = $ 25} & \multicolumn{2}{c}{$k = $ 50} & \multicolumn{2}{c}{$k = $ 100} & \multicolumn{2}{c}{$k = $ 200} & \multicolumn{2}{c}{$k = $ 25} & \multicolumn{2}{c}{$k = $ 50} & \multicolumn{2}{c}{$k = $ 100} & \multicolumn{2}{c}{$k = $ 200} \\
    \cmidrule(lr){3-4}\cmidrule(lr){5-6}\cmidrule(lr){7-8}\cmidrule(lr){9-10}\cmidrule(lr){11-12}\cmidrule(lr){13-14}\cmidrule(lr){15-16}\cmidrule(lr){17-18}
     &  & $\bm{\Theta}$ & $\textbf{B}$ & $\bm{\Theta}$ & $\textbf{B}$ & $\bm{\Theta}$ & $\textbf{B}$ & $\bm{\Theta}$ & $\textbf{B}$ & $\bm{\Theta}$ & $\textbf{B}$ & $\bm{\Theta}$ & $\textbf{B}$ & $\bm{\Theta}$ & $\textbf{B}$ & $\bm{\Theta}$ & $\textbf{B}$ \\
    
    \midrule
    \multirow[c]{5}{*}{Bills} & \mallet{} & \textbf{0.78} & \textbf{0.27} & \textbf{0.74} & \textbf{0.28} & \textbf{0.74} & \textbf{0.32} & 0.79 & \textbf{0.41} & \textbf{0.79} & \textbf{0.29} & \textbf{0.80} & \textbf{0.33} & \textbf{0.77} & \textbf{0.35} & \underline{0.76} & \textbf{0.39} \\
     & \scholar{} & 0
    .88 & 0.63 & 0.82 & 0.56 & \underline{0.76} & 0.50 & 0.78 & 0.57 & 0.86 & 0.82 & 0.85 & 0.76 & 0.83 & 0.71 & 0.84 & 0.71 \\
     & \scholarkd{} & 0.91 & 0.67 & 0.89 & 0.59 & 0.87 & 0.64 & 0.83 & 0.65 & 0.90 & 0.75 & 0.88 & 0.69 & 0.85 & 0.67 & 0.88 & 0.70 \\
     & \dvae{} & 0.97 & 0.62 & 0.97 & 0.77 & 0.96 & 0.73 & 0.96 & 0.76 & 0.97 & 0.75 & 0.97 & 0.81 & 0.97 & 0.84 & 0.95 & 0.83 \\
     & \ctm{} & \underline{0.79} & 0.43 & 0.80 & 0.51 & 0.76 & 0.54 & \textbf{0.74} & 0.58 & \underline{0.81} & 0.44 & \underline{0.83} & 0.55 & 0.81 & 0.60 & \textbf{0.76} & 0.65 \\
    \cline{1-18}
    \multirow[c]{5}{*}{Wiki} & \mallet{} & \textbf{0.70} & \textbf{0.22} & \textbf{0.69} & \textbf{0.29} & \textbf{0.62} & \textbf{0.30} & \textbf{0.67} & \textbf{0.37} & \textbf{0.71} & \textbf{0.26} & \textbf{0.70} & \textbf{0.32} & \textbf{0.66} & \textbf{0.34} & \textbf{0.70} & \textbf{0.39} \\
     & \scholar{} & 0.82 & 0.49 & 0.73 & 0.38 & 0.80 & 0.45 & 0.83 & 0.52 & 0.84 & 0.66 & 0.77 & 0.54 & 0.77 & 0.67 & 0.79 & 0.61 \\
     & \scholarkd{} & 0.83 & 0.47 & 0.80 & 0.47 & 0.86 & 0.52 & 0.83 & 0.42 & 0.88 & 0.65 & 0.84 & 0.60 & 0.86 & 0.64 & 0.89 & 0.60 \\
     & \dvae{} & 0.92 & 0.46 & 0.92 & 0.55 & 0.91 & 0.54 & 0.92 & 0.67 & 0.92 & 0.67 & 0.92 & 0.63 & 0.90 & 0.76 & 0.87 & 0.72 \\
     & \ctm{} & 0.76 & 0.42 & 0.73 & 0.39 & 0.73 & 0.46 & 0.72 & 0.51 & 0.76 & 0.39 & 0.76 & 0.43 & 0.76 & 0.50 & 0.73 & 0.56 \\
    \cline{1-18}
    \bottomrule
    \end{tabular}
    }
    \caption{Stability for topic-word $\textbf{B}$ and document-topic $\bm{\Theta}$ estimates, over 10 runs. Smallest per-column values are \textbf{bolded} and are sig. smaller than unbolded values (2-sided t-test, $p<0.05$); \underline{underlined} values have $p>0.05$.}
    \label{tab:stability-results}
\end{table*}

\begin{table*}[ht]
\label{tab:cluster-results}
\centering
\resizebox{.85\textwidth}{!}{
\begin{tabular}{llrrrrrrrrrrrr}
\toprule
 &  & \multicolumn{3}{c}{$k = $ 25} & \multicolumn{3}{c}{$k = $ 50} & \multicolumn{3}{c}{$k = $ 100} & \multicolumn{3}{c}{$k = $ 200}\\
\cmidrule(lr){3-5}\cmidrule(lr){6-8}\cmidrule(lr){9-11}\cmidrule(lr){12-14}
 &  & ARI & NMI & $P_1$ & ARI & NMI & $P_1$ & ARI & NMI & $P_1$ & ARI & NMI & $P_1$ \\
\midrule
\multirow[c]{5}{*}{\shortstack[l]{Bills \\ $\ell=114$}} & \mallet{} & \textbf{0.30} & \textbf{0.45} & \textbf{0.46} & \textbf{0.34} & \textbf{0.48} & \textbf{0.47} & \textbf{0.32} & \textbf{0.50} & \textbf{0.43} & 0.22 & \textbf{0.50} & \underline{0.35} \\
 & \scholar{} & 0.12 & 0.28 & 0.27 & 0.19 & 0.40 & 0.34 & 0.15 & 0.40 & 0.29 & 0.12 & 0.39 & 0.25 \\
 & \scholarkd{} & 0.11 & 0.28 & 0.27 & 0.16 & 0.37 & 0.35 & 0.14 & 0.41 & 0.33 & 0.11 & 0.38 & 0.25 \\
 & \dvae{} & \underline{0.26} & \underline{0.45} & \underline{0.44} & 0.24 & \underline{0.43} & \underline{0.40} & 0.24 & 0.46 & \underline{0.40} & \textbf{0.24} & 0.46 & \textbf{0.38} \\
 & \ctm{} & 0.21 & 0.40 & 0.38 & 0.26 & 0.45 & 0.41 & 0.25 & 0.48 & 0.39 & 0.19 & \underline{0.49} & \underline{0.34} \\
\cline{1-14}
\multirow[c]{5}{*}{\shortstack[l]{Wiki \\ $\ell=279$}} & \mallet{} & \textbf{0.23} & \textbf{0.65} & \textbf{0.41} & \textbf{0.32} & \textbf{0.69} & \textbf{0.50} & \textbf{0.37} & \textbf{0.71} & \textbf{0.53} & \underline{0.32} & \textbf{0.70} & \underline{0.48} \\
 & \scholar{} & \underline{0.21} & 0.61 & 0.38 & \underline{0.31} & 0.68 & 0.48 & \underline{0.34} & \underline{0.69} & \underline{0.50} & \underline{0.29} & \underline{0.68} & \underline{0.44} \\
 & \scholarkd{} & 0.19 & 0.61 & 0.37 & 0.26 & 0.65 & 0.43 & 0.29 & 0.65 & 0.44 & 0.22 & 0.62 & 0.36 \\
 & \dvae{} & \underline{0.22} & \underline{0.64} & \underline{0.39} & \underline{0.30} & \underline{0.68} & \underline{0.48} & 0.27 & 0.65 & 0.45 & \underline{0.30} & \underline{0.68} & \textbf{0.49} \\
 & \ctm{} & 0.21 & 0.60 & 0.36 & 0.27 & 0.64 & 0.43 & 0.31 & 0.67 & 0.46 & \textbf{0.34} & \underline{0.69} & \underline{0.47} \\
\cline{1-14}
\bottomrule
\end{tabular}
}
\caption{
    Average alignment metrics across 10 runs, 
    measured against gold labels at the lowest hierarchy level, $|V| = 15,000$. 
    Largest values in each column are \textbf{bolded}, which are
    significantly greater than unbolded values in a two-sided t-test ($p<0.05$);
    \underline{underlined} values have $p>0.05$.}
\label{tab:cluster-results}
\end{table*}

Recall that measuring \emph{stability} is motivated by intra-coder reliability in content analysis: producing the same result every time increases confidence that the analysis reflects actual latent structure in the data.
\mallet{} is significantly more stable than other models across the vast majority of contexts, often by a large margin (Table~\ref{tab:stability-results}). 
Most striking are the topic-word distributions $\textbf{B}$: none of the neural models even approach its consistent level of stability.
%
%
\ctm{} sometimes achieves comparable stability for $\bm{\Theta}$; this may be due to its use of pretrained document embeddings, which are transformed in order to parameterize the estimate. 

Recall also that \emph{alignment} is motivated by \emph{inter}-coder reliability in content analysis: is the model, in the role of analyst, agreeing with human-derived categorization for the data?
\mallet{} shows strong consistency in providing the numerically best alignment with human categorization across datasets (Table~\ref{tab:cluster-results}).\footnote{The distribution of scores across runs and results for other contexts are shown in Appendix~\ref{sec:app:alignment}.} Among neural models, \abr{d-vae} and \abr{\scholar{}} sometimes achieve statistically indistinguishable performance, but they do not approach \mallet{}'s consistency across datasets, number of topics, and metrics.

\ignore{
  Overwhelmingly, measurements of alignment to do not vary widely across models (\cref{tab:cluster-results}).
However, there are broad trends: \mallet{} tends to have significantly larger values than all but \abr{d-vae} and, for the wiki dataset \abr{\scholar{}}. \todo{elaborate more}
\todo{Revise previous paragraphs; above was just to get something down}
}


Now, we argued in \S\ref{sec:background:stability} that practitioners do not have access to optimal hyperparameters for a given model, because what is optimal will depend on the dataset, number of topics, preprocessing, and other experimental decisions.
The above results show that model estimates can be very sensitive to different hyperparameter settings and they clearly favor \mallet{} on our metrics. However, in many real-world scenarios, a practitioner may simply rely on some ``default'' settings.
We therefore also evaluate models for \emph{fixed} hyperparameters using reasonable default values.\footnote{We thank an anonymous reviewer for pointing this out and suggesting the additional experimentation.}

To generate the defaults, for each dataset and model we find the hyperparameter settings that yield the best alignment performance across experimental contexts (vocabulary size, number of topics, alignment metric, and label hierarchy).
Specifically, within each context, we first rescale the alignment metric values over the 10 runs for that model to avoid differences in metric values; we then select the hyperparameters which have the largest average values across all contexts, for a given dataset. 
Finally, to approximate a common use case and to avoid overfitting to the dataset, we use the hyperparameters obtained from one dataset to train models on the \emph{other} dataset (e.g., we select defaults based on the \texttt{Bill} alignment metrics and set those for new models run on \texttt{Wiki}; defaults in Appendix~\ref{sec:app:hparams}).

Results are in Appendix~\ref{app:fixed-hparam}. Unsurprisingly, fixing ``good default'' hyperparameters for the neural models improves their stability and alignment. In particular, \dvae{} has competitive alignment metrics in the $|V|=5k$ case, although it is hampered by its relatively poor stability. \mallet{}'s stability is marginally affected: while it is no longer as consistently dominant, it remains more stable and better-aligned in the majority of contexts.

\section{A close reading of model stability}
\label{sec:closereading}

Table~\ref{tab:top-word-comparison} illustrates corresponding versions of a topic from different runs of \dvae{} and \mallet{}.
For a given context (here, $K=50, |V|=15,000$), we collect the topic-word estimates $\textbf{B}$ across the 10 runs for each of the two models, each run using a different set of randomly-selected hyperparameters. One weather-related topic across runs was chosen manually as the ``base'' run, and then the corresponding topics in the other nine runs for the same model were ranked by their RBO distance to that topic. The nearest, median, and most-distant topics in that ranking, shown in the table, therefore capture the range of variation across different hyperparameter settings.
%
%
%
%
%
%
%
 
 It is immediately clear that even the nearest topic for \dvae{} has fewer words in common with the base topic, compared to \mallet{}.
And as distance increases, the top words for \mallet{} stay consistent, whereas those for \dvae{} change dramatically, even if they relate to the same \underline{weather} concept. \ignore{ (further note that some concepts may not appear in all runs).}
Note that in this example, consistent with anecdotal reports from other practitioners and our own experience, the neural model tends toward less frequent or more specific words.  The idea that neural models may be capturing topics that are in some sense narrower, with instability leading to different such topics in each run, leads directly to the idea that a cross-run ensemble might be expected perform better than the individual runs---which is important in the absence of a reliable automated method for optimizing hyperparameters.

\ignore{
  Some of the words appearing in \dvae{} are highly specific, potentially affecting interpretability.
  }

\section{Ensembling estimates}\label{sec:ensemble}

\begin{figure*}
	\centerline{
	\includegraphics[width=\textwidth]{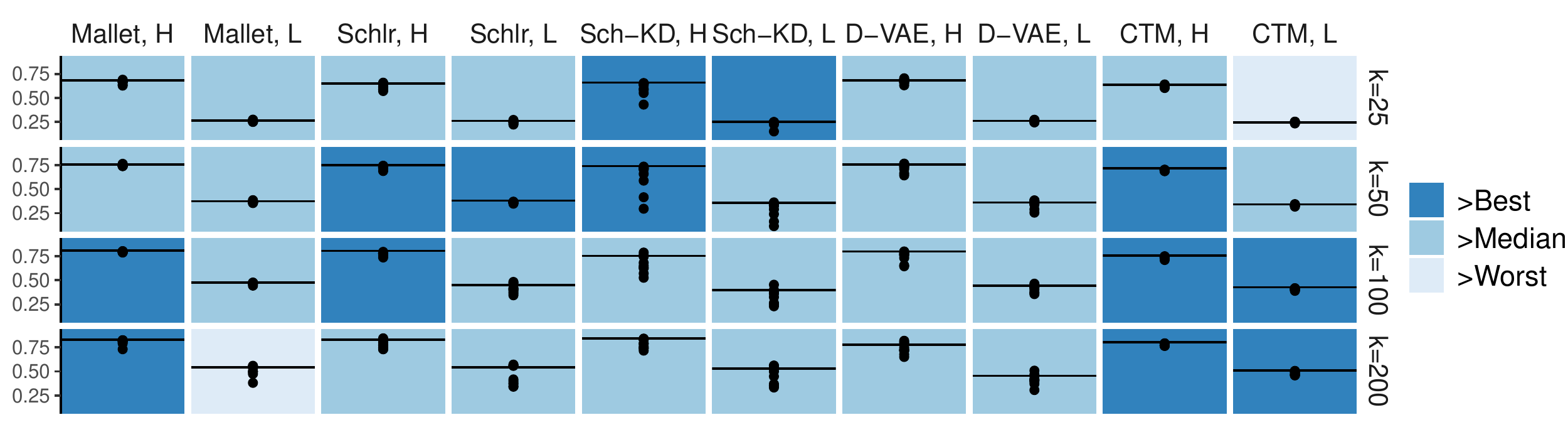}}
\caption{
    Ensembling performance on alignment (purity metric, \S~\ref{sec:background:cluster}) for the \texttt{Wiki} dataset.
    Each box represents a context: the columns identify model type and the label granularity used in evaluation (e.g. top left is \mallet{} with \underline{\bf H}igh-level categories), and the rows correspond to different values $K$ for the number of topics.
    Dots are alignment scores for individual runs; the horizontal line is the alignment score for that \emph{ensemble} of runs using our method. Shading indicates when the ensemble method has beaten the score of the best individual run (darkest), the median (middle), or has outperformed the worst individual run (lightest). The ensemble is typically in the top quartile of the component runs. Ensembles virtually always outdo the median, and frequently outperform the best individual run.  
}
\label{fig:ensemble}
\end{figure*}

We have highlighted lack of stability as a serious problem for neural topic models, but neural models can also have desirable properties.
How can we increase the odds of obtaining a good neural topic model in the face of extreme variation? 
The distance metrics we use to measure instability offer one solution: clustering to aggregate similar estimates over runs to form an \emph{ensemble}. We adopt an approach similar to prior work \cite{miller-mccoy-2017-topic,mantyla2018measuring}, going further by accounting for the document-word estimates $\bm{\Theta}$ and by evaluating ensembles' alignment against human categorization.
Specifically, we concatenate run estimates over the $m$ runs $\bar{\textbf{B}}=\left[\textbf{B}^{(i)}\right]^m_{i}$ and $\bar{\bm{\Theta}}=\left[\bm{\Theta}^{\top(i)}\right]^m_{i}$, where each row in the concatenated matrix is a topic. We then compute pairwise distances between topics, $\mathcal{D}(\bar{\textbf{B}})$ and $\mathcal{D}(\bar{\bm{\Theta}})$, and cluster based on a linear interpolation of the two distances, $\lambda\mathcal{D}(\bar{\textbf{B}}) + (1-\lambda)\mathcal{D}(\bar{\bm{\Theta}})$, where $\lambda$ is a hyperparameter.
The estimate of each topic $k$ from each run $i$, $\langle \bm{\theta}_k^{\top(i)}, \bm{\beta}_k^{(i)} \rangle$, is assigned to a cluster, and to infer new document-topic or topic-word scores for the ensemble, we take the element-wise mean over the estimates assigned to each cluster.\footnote{We experimented with $k$-medoids \citep{lloyd1957least,kaufman} and hierarchical agglomerative clustering \citep{day1984efficient}, and also varied the distance metric --- either RBO (\S\ref{sec:background:stability}) or the Average Jaccard score \cite{greene2014many}. Both metrics are used to measure the distance between topics in prior work \citep{mantyla2018measuring,greene2014many}. Clustering algorithms were implemented using scikit-learn \citep{pedregosa2011scikit}.}

To evaluate this method, we compare the alignment score of the ensemble (\S\ref{sec:background:cluster}) combining the $m=10$ runs, versus  the alignment score of each individual member run.  We do so across each of the 400 contexts (model, dataset, $K$, high versus low label granularity, and metric).
\ignore{
To evaluate this method, we compare the ensemble's alignment score (\S\ref{sec:background:cluster}) to the scores of its $m=10$ individual members across each of the 400 contexts (model, dataset, $K$, label granularity, and metric).}
Figure~\ref{fig:ensemble} illustrates a summary of results for the purity alignment metric on the Wikipedia dataset.
Across the full range of our experimentation, the ensemble improves on the median member in 97\% of all the contexts, and it is \emph{always} better than the worst member (full results in Appendix~\ref{sec:app:ensemble}).

\ignore{Agglomerative clustering can induce a number of topics using a distance threshold, allowing for user specification of cluster granularity based on their assessment of word coherence \todo{possibly cut}.}

\section{Conclusions}\label{sec:conclusion}


A tool can be considered broken when it doesn't work well for its intended use. In this paper we have focused on a widespread use case for topic models, their application in text content analysis; we have carefully motivated criteria for measuring the extent to which a topic model is serving those needs; and we have demonstrated through comprehensive and replicable experimentation that, when measured on those criteria, recent and representative neural topic models fail to improve on the classical implementation of \lda{} in \mallet{}.  In particular, \mallet{} is much more stable, reducing concerns from the content analysis perspective that different runs could yield very different codesets. Equally important, across the vast majority of contexts, its discovered categories are reliable as measured via alignment with ground-truth human categories. For people seeking to use topic modeling in content analysis, therefore, \mallet{} may still be the best available tool for the job.

That said, there are still good reasons to investigate neural topic models. Foremost among these is the fact that they can benefit from pretraining on vast, general samples of language \citep[e.g.][]{hoyle-etal-2020-improving,bianchi-etal-2021-pre,feng2022context}. Neural realizations of topic models can also be integrated smoothly for joint modeling within larger neural architectures \citep[e.g.][]{lau-etal-2017-topically,wang-etal-2019-topic,wang-etal-2020-friendly}, and hold the promise of being more straightforward to use multilingually \citep[e.g.][]{wu2020learning,bianchi-etal-2021-cross,mueller-dredze-2021-fine} or multimodally \citep[e.g.][]{zheng2015deep}.\footnote{See \citet{ijcai2021-638} for more potential advantages of neural topic modeling.} 
We therefore introduced one possible way to address the shortcomings we identified using a straightforward ensemble technique.

Perhaps the most important take-away we would suggest is that \emph{development} of new topic models---indeed, of all NLP models---should be done with \emph{use cases} firmly in mind.  Some models are enabling technologies, without a direct user-facing purpose, and others are intended to produce results directly for human consumption.  But whatever the goal, the driving question for methodological development and evaluation should \emph{not} be how to demonstrate an improvement in ``state of the art'', it should be why the model is being created in the first place and what measurements will demonstrate improved performance for that intended purpose.

\section*{Limitations}

Our studies used only English datasets, while topic modeling has been used to characterize texts in many languages. While \emph{theoretically} we see no reason why our results and findings should not generalize beyond the English language, \emph{empirical} generalizability across languages remains to be determined.

Our method for measuring alignment of model-induced categories with human-determined categories relies on ground-truth human labels, potentially limiting its broader applicability. In addition, the categories in the Wikipedia data were not, to our knowledge, produced via a traditional human content analysis process. We are currently designing a follow-up study in which human subject matter experts perform traditional content analysis from scratch on the same dataset used for topic modeling, in order to provide a head-to-head comparison between automated and traditional methods and to establish human upper bounds on inter-coder reliability.

Our literature review of topic modeling use cases was not a formal systematic review \citep{moher2009preferred}. It relied on Semantic~Scholar's content and its discipline categorization, and potentially excluded papers in computer science that were about the use of topic models rather than method development. It seems clear that text content analysis \emph{a} dominant use case for topic modeling, if not \emph{the} dominant use case. In the social sciences, we also note frequent use of the Structural Topic Model \cite{roberts2014structural} which, like \scholar{}, can incorporate metadata into model estimation---we leave an evaluation of this use case to future work.
\ignore{
\section{Introduction}

\section{Definitions}

\paragraph{Topic Assignment.} Topic models are often used to \emph{assign} documents a label based on their estimated document-topic distribution,  $\ell_d = f(\theta_d)$. In practice, $f(\cdot)$ could be an argmax, a threshold, or computed based on the values of other $\theta_{d'}$ or the topic-word distributions $\bm{\beta}$. Note that it is possible for a document to fail to be assigned, $\ell_d = \emptyset$.

\paragraph{Junk topic.} A junk topic is a topic whose topic-word distribution and assigned documents do not correspond to a consistent, coherent latent concept.

\paragraph{Corpus Coverage.} Directly related to the question of assignment is that of \emph{coverage}. Once we have a labeling over all documents, we can compute the number of documents with assignments to non-junk topics as a proportion of the total corpus.

\paragraph{Supervised Coverage.} This is a measurement of the extent to which the assigned labelings agree with external labelings. 

\paragraph{Stability.} Stability is a measurement of how consistent the topic model estimates are for the same corpus.

\section{Measuring Stability}
Intuitively, estimates from a model on the same set of data should be similar over multiple runs: if there is meaningful latent structure to the data, then we expect that models consistently uncover that structure. Ideally, estimates should remain stable both within and across hyperparameter settings. Here, we present two ways of measuring model stability, the first uses topic-word estimates $\bm{\beta}$ and the second the document-topic estimates $\bm{\Theta}$.

\paragraph{Topic-word stability.}
To measure topic word stability, we collect a set of $\bm{\beta}_i, i=1\ldots n$ estimates from $n$ models on the same dataset. For each pair of $n \choose 2$ runs, we compute the pairwise Jensen-Shannon distance $d$ between all $k$ topics in each run. We then minimize the total bipartite matching distance with a modified Jonker-Volgenant algorithm \cite{crouse2016,hoyle-etal-2020-improving}. If the average total distances are smaller, this implies a more stable model. 


To measure document-topic stability, we have

\begin{align}
    \text{AS}\left(\bm{\Theta}\right) = \frac{1}{l R} \left| \bigcup_{i \in R}  \text{argsort}_{1:l}\left(\mathbf{B}_{\theta_{k^*}^{(i)}}\right) \right|
\end{align}

\section{Coverage}

\section{Human evaluations}

\section{Related Work}

\pg{first just noting all the things not covered in the rest of the paper we may want to have, and then for each of the points below, we have brief paragraphs with the related work -- any other cluster of work that should be in this related work section?}

\begin{itemize}

\item Stability and coverage assessment in other topic modeling work - mostly done for LDA so we note those works and briefly comment on what they do (like try to use it to tune settings like number of topics for LDA), we note the one or two neural topic modeling works that did look at these things, and contextualize our work. And the meta-analysis on NTM can be summarized here pointing out how basically most NTM work has ignored looking at these things?

\item Brief background on neural topic models? with autoencoders? (Methodology only, as some use cases and good reasons to investigate neural topic models and examples are already there in conclusion section).

\item Perhaps a brief note on Computer Assisted Content Analysis, where topic models are a part of that assistance. May not need this. One distinction between this and the meta-analysis on LDA is that here we could talk about cases where improving content analysis using topic models was the goal of the work? 

\end{itemize}

}
\section{Acknowledgements}
This material is based upon work supported in part by the National Science Foundation under Grants 2031736 and 2008761 and by Amazon. We thank  David Mimno and Xanda Schofield for their input on earlier drafts, as well as our anonymous reviewers for their helpful comments. 

\bibliography{anthology,custom}
\bibliographystyle{acl_natbib}

\newpage
\appendix

\section{Appendix}\label{sec:app}

\begin{table*}[ht]
\resizebox{\textwidth}{!}{
\begin{tabular}{llrrrrrrrrrrrrrrrr}
\toprule
 &  & \multicolumn{8}{c}{$|V| = $ 5k} & \multicolumn{8}{c}{$|V| = $ 15k} \\
\cmidrule(lr){3-10}\cmidrule(lr){11-18}
 &  & \multicolumn{2}{c}{$k = $ 25} & \multicolumn{2}{c}{$k = $ 50} & \multicolumn{2}{c}{$k = $ 100} & \multicolumn{2}{c}{$k = $ 200} & \multicolumn{2}{c}{$k = $ 25} & \multicolumn{2}{c}{$k = $ 50} & \multicolumn{2}{c}{$k = $ 100} & \multicolumn{2}{c}{$k = $ 200} \\
\cmidrule(lr){3-4}\cmidrule(lr){5-6}\cmidrule(lr){7-8}\cmidrule(lr){9-10}\cmidrule(lr){11-12}\cmidrule(lr){13-14}\cmidrule(lr){15-16}\cmidrule(lr){17-18}
 &  & $\bm{\Theta}$ & $\textbf{B}$ & $\bm{\Theta}$ & $\textbf{B}$ & $\bm{\Theta}$ & $\textbf{B}$ & $\bm{\Theta}$ & $\textbf{B}$ & $\bm{\Theta}$ & $\textbf{B}$ & $\bm{\Theta}$ & $\textbf{B}$ & $\bm{\Theta}$ & $\textbf{B}$ & $\bm{\Theta}$ & $\textbf{B}$ \\

\midrule
\multirow[c]{5}{*}{Bills} & \mallet{} & 0.76 & \textbf{0.26} & 0.75 & \textbf{0.29} & 0.73 & \textbf{0.30} & 0.70 & 0.36 & \textbf{0.73} & \textbf{0.26} & 0.74 & \textbf{0.30} & 0.72 & 0.34 & \textbf{0.70} & 0.37 \\
 & \scholar{} & \textbf{0.72} & 0.42 & \textbf{0.67} & 0.39 & \textbf{0.67} & 0.38 & \textbf{0.68} & 0.38 & 0.78 & 0.52 & \textbf{0.73} & 0.47 & \textbf{0.70} & 0.44 & 0.71 & 0.43 \\
 & \scholarkd{} & 0.84 & 0.43 & 0.78 & 0.38 & 0.70 & \underline{0.31} & 0.69 & \textbf{0.29} & 0.83 & 0.46 & 0.80 & 0.38 & 0.74 & \textbf{0.31} & 0.72 & \textbf{0.29} \\
 & \dvae{} & 0.91 & 0.34 & 0.86 & 0.54 & 0.85 & 0.68 & 0.74 & 0.81 & 0.95 & 0.40 & 0.90 & 0.57 & 0.90 & 0.71 & 0.91 & 0.79 \\
 & \ctm{} & 0.78 & 0.43 & 0.77 & 0.47 & 0.73 & 0.51 & 0.70 & 0.52 & 0.80 & 0.44 & 0.81 & 0.53 & 0.79 & 0.60 & 0.73 & 0.62 \\
\cline{1-18}
\multirow[c]{5}{*}{Wiki} & \mallet{} & \textbf{0.70} & \textbf{0.22} & \textbf{0.60} & \underline{0.26} & \textbf{0.56} & 0.28 & 0.54 & 0.31 & \textbf{0.69} & \textbf{0.26} & \textbf{0.62} & \textbf{0.29} & \textbf{0.55} & \textbf{0.29} & \textbf{0.50} & \textbf{0.31} \\
 & \scholar{} & 0.77 & 0.39 & 0.66 & 0.31 & 0.62 & 0.33 & 0.59 & 0.32 & 0.75 & 0.45 & 0.69 & 0.38 & 0.65 & 0.38 & 0.60 & 0.37 \\
 & \scholarkd{} & 0.77 & 0.41 & 0.66 & 0.33 & 0.62 & 0.33 & \textbf{0.53} & \textbf{0.28} & 0.78 & 0.52 & 0.70 & 0.44 & 0.62 & 0.38 & 0.54 & 0.34 \\
 & \dvae{} & 0.93 & 0.24 & 0.88 & \textbf{0.25} & 0.83 & \textbf{0.26} & 0.82 & 0.29 & 0.95 & 0.28 & 0.90 & \underline{0.30} & 0.83 & 0.32 & 0.78 & 0.33 \\
 & \ctm{} & 0.75 & 0.39 & 0.73 & 0.39 & 0.72 & 0.45 & 0.70 & 0.51 & 0.77 & 0.42 & 0.74 & 0.41 & 0.75 & 0.51 & 0.73 & 0.57 \\
\cline{1-18}
\bottomrule
\end{tabular}
}
\caption{
Stability for topic-word $\textbf{B}$ and document-topic $\bm{\Theta}$ estimates, across 10 seeds, for \emph{fixed hyperparameters}. 
Smallest per-column values are \textbf{bolded} and are
 sig. smaller than unbolded values (2-sided t-test, $p<0.05$);
 \underline{underlined} values have $p>0.05$.
}
\label{tab:stability-results-by-seed}
\end{table*}

\begin{table*}[ht]
\centering
\resizebox{.85\textwidth}{!}{
\begin{tabular}{llrrrrrrrrrrrr}
\toprule
 &  & \multicolumn{3}{c}{$k = $ 25} & \multicolumn{3}{c}{$k = $ 50} & \multicolumn{3}{c}{$k = $ 100} & \multicolumn{3}{c}{$k = $ 200}\\
\cmidrule(lr){3-5}\cmidrule(lr){6-8}\cmidrule(lr){9-11}\cmidrule(lr){12-14}
 & & ARI & NMI & $P_1$ & ARI & NMI & $P_1$ & ARI & NMI & $P_1$ & ARI & NMI & $P_1$ \\
\midrule
\multirow[c]{5}{*}{\shortstack[l]{Bills \\ $\ell=114$}} & \mallet{} & \textbf{0.30} & \textbf{0.46} & \textbf{0.47} & \textbf{0.35} & \textbf{0.49} & \textbf{0.48} & \textbf{0.34} & \underline{0.51} & \textbf{0.45} & \textbf{0.22} & 0.51 & 0.37 \\
 & \scholar{} & 0.25 & 0.45 & 0.42 & 0.25 & \underline{0.48} & 0.43 & 0.21 & \textbf{0.51} & 0.40 & 0.15 & \textbf{0.52} & 0.34 \\
 & \scholarkd{} & 0.24 & 0.42 & 0.40 & 0.23 & 0.46 & 0.43 & 0.20 & 0.49 & 0.40 & 0.13 & 0.49 & 0.34 \\
 & \dvae{} & 0.26 & 0.45 & 0.45 & 0.21 & 0.45 & 0.45 & 0.10 & 0.43 & 0.42 & 0.04 & 0.39 & \textbf{0.41} \\
 & \ctm{} & 0.23 & 0.40 & 0.39 & 0.27 & 0.46 & 0.43 & 0.24 & 0.48 & 0.40 & 0.18 & 0.49 & 0.34 \\
\cline{1-14}
\multirow[c]{5}{*}{\shortstack[l]{Wiki \\ $\ell=279$}} & \mallet{} & 0.23 & 0.65 & \underline{0.41} & 0.32 & \underline{0.70} & \textbf{0.50} & 0.39 & \textbf{0.73} & \textbf{0.56} & \textbf{0.39} & 0.74 & \textbf{0.56} \\
 & \scholar{} & 0.22 & 0.62 & 0.39 & \textbf{0.33} & 0.68 & 0.49 & \textbf{0.39} & 0.72 & 0.54 & \underline{0.38} & \textbf{0.74} & 0.54 \\
 & \scholarkd{} & 0.20 & 0.61 & 0.37 & 0.30 & 0.67 & 0.47 & \underline{0.39} & 0.71 & 0.53 & \underline{0.39} & 0.74 & 0.54 \\
 & \dvae{} & \textbf{0.24} & \textbf{0.66} & \textbf{0.41} & 0.32 & \textbf{0.70} & \underline{0.50} & 0.36 & 0.72 & 0.54 & 0.36 & 0.72 & 0.54 \\
 & \ctm{} & 0.21 & 0.60 & 0.36 & 0.28 & 0.65 & 0.44 & 0.32 & 0.67 & 0.47 & 0.35 & 0.70 & 0.48 \\
\cline{1-14}
\bottomrule
\end{tabular}
}
\caption{
Average alignment metrics across 10 seeds, for \emph{fixed hyperparameters}. 
Measured against gold labels at the lowest hierarchy level, $|V| = 15,000$.
Largest values in each column are \textbf{bolded}, which are
 significantly greater than unbolded values in a two-sided t-test ($p<0.05$);
 \underline{underlined} values have $p>0.05$.
}
\label{tab:cluster-results-by-seed}
\end{table*}

\begin{table*}[ht]
    \centering
    
    \begin{subtable}{\textwidth}
    \centering
    \resizebox{.85\textwidth}{!}{
    \begin{tabular}{llrrrrrrrrrrrr}
    \toprule
     &  & \multicolumn{3}{c}{$k = $ 25} & \multicolumn{3}{c}{$k = $ 50} & \multicolumn{3}{c}{$k = $ 100} & \multicolumn{3}{c}{$k = $ 200}\\
    \cmidrule(lr){3-5}\cmidrule(lr){6-8}\cmidrule(lr){9-11}\cmidrule(lr){12-14}
     & & ARI & NMI & $P_1$ & ARI & NMI & $P_1$ & ARI & NMI & $P_1$ & ARI & NMI & $P_1$ \\
    \midrule
    \multirow[c]{5}{*}{\shortstack[l]{Bills \\ $\ell=114$}} & \mallet{} & \textbf{0.17} & \textbf{0.34} & \textbf{0.37} & \textbf{0.17} & \textbf{0.37} & \textbf{0.37} & \textbf{0.17} & \textbf{0.40} & \textbf{0.37} & \textbf{0.15} & \textbf{0.41} & \textbf{0.35} \\
     & \scholar{} & 0.06 & 0.18 & 0.22 & 0.09 & 0.27 & 0.26 & 0.07 & 0.27 & 0.23 & 0.05 & 0.26 & 0.19 \\
     & \scholarkd{} & 0.06 & 0.19 & 0.24 & 0.07 & 0.25 & 0.27 & 0.06 & 0.27 & 0.26 & 0.04 & 0.26 & 0.19 \\
     & \dvae{} & 0.13 & 0.30 & 0.34 & 0.11 & 0.29 & 0.32 & 0.11 & 0.32 & 0.32 & 0.10 & 0.34 & 0.31 \\
     & \ctm{} & 0.11 & 0.28 & 0.31 & 0.12 & 0.32 & 0.31 & 0.10 & 0.35 & 0.29 & 0.08 & 0.37 & 0.26 \\
    \cline{1-14}
    \multirow[c]{5}{*}{\shortstack[l]{Wiki \\ $\ell=279$}} & \mallet{} & \textbf{0.34} & \textbf{0.65} & \textbf{0.53} & \textbf{0.37} & \textbf{0.66} & \textbf{0.53} & \textbf{0.37} & \textbf{0.68} & \textbf{0.53} & \textbf{0.33} & \textbf{0.68} & \textbf{0.51} \\
     & \scholar{} & 0.29 & 0.61 & 0.48 & 0.32 & 0.64 & 0.51 & 0.30 & 0.63 & 0.47 & 0.24 & 0.61 & 0.41 \\
     & \scholarkd{} & 0.28 & 0.61 & 0.48 & 0.31 & 0.63 & 0.49 & 0.28 & 0.63 & 0.46 & 0.19 & 0.59 & 0.35 \\
     & \dvae{} & 0.32 & 0.64 & 0.51 & \underline{0.35} & 0.66 & \underline{0.52} & 0.28 & 0.62 & 0.48 & \underline{0.31} & 0.64 & \underline{0.48} \\
     & \ctm{} & 0.32 & 0.61 & 0.49 & 0.32 & 0.63 & 0.48 & 0.29 & 0.64 & 0.46 & 0.27 & 0.65 & 0.45 \\
    \cline{1-14}
    \bottomrule
    \end{tabular}
    }
    \caption{Random hyperparameters}
    \end{subtable}

    \begin{subtable}{\textwidth}
        \centering
        \resizebox{.85\textwidth}{!}{
        \begin{tabular}{llrrrrrrrrrrrr}
        \toprule
         &  & \multicolumn{3}{c}{$k = $ 25} & \multicolumn{3}{c}{$k = $ 50} & \multicolumn{3}{c}{$k = $ 100} & \multicolumn{3}{c}{$k = $ 200}\\
        \cmidrule(lr){3-5}\cmidrule(lr){6-8}\cmidrule(lr){9-11}\cmidrule(lr){12-14}
         & & ARI & NMI & $P_1$ & ARI & NMI & $P_1$ & ARI & NMI & $P_1$ & ARI & NMI & $P_1$ \\
        \midrule
        \multirow[c]{5}{*}{\shortstack[l]{Bills \\ $\ell=114$}} & \mallet{} & \textbf{0.17} & \textbf{0.34} & \textbf{0.37} & \textbf{0.17} & \textbf{0.37} & \textbf{0.38} & \textbf{0.17} & \textbf{0.40} & \textbf{0.38} & \textbf{0.15} & \textbf{0.42} & \textbf{0.36} \\
         & \scholar{} & 0.14 & 0.32 & 0.34 & 0.13 & 0.34 & 0.34 & 0.14 & 0.38 & 0.34 & 0.12 & 0.40 & 0.32 \\
         & \scholarkd{} & 0.10 & 0.27 & 0.31 & 0.09 & 0.29 & 0.32 & 0.08 & 0.33 & 0.30 & 0.06 & 0.32 & 0.31 \\
         & \dvae{} & 0.11 & 0.27 & 0.32 & 0.05 & 0.26 & 0.31 & 0.03 & 0.25 & 0.32 & 0.01 & 0.21 & 0.34 \\
         & \ctm{} & 0.11 & 0.28 & 0.31 & 0.12 & 0.32 & 0.31 & 0.10 & 0.34 & 0.29 & 0.07 & 0.36 & 0.26 \\
        \cline{1-14}
        \multirow[c]{5}{*}{\shortstack[l]{Wiki \\ $\ell=279$}} & \mallet{} & 0.33 & \underline{0.65} & \underline{0.52} & 0.37 & 0.66 & 0.54 & \textbf{0.38} & \underline{0.68} & \textbf{0.54} & \textbf{0.34} & 0.69 & \textbf{0.52} \\
         & \scholar{} & 0.31 & 0.62 & 0.50 & 0.33 & 0.64 & 0.50 & 0.34 & 0.67 & 0.51 & 0.32 & 0.68 & 0.49 \\
         & \scholarkd{} & 0.30 & 0.63 & 0.49 & \textbf{0.38} & \textbf{0.67} & \textbf{0.54} & \underline{0.37} & \textbf{0.69} & 0.52 & \underline{0.34} & \textbf{0.70} & 0.51 \\
         & \dvae{} & \textbf{0.34} & \textbf{0.65} & \textbf{0.52} & \underline{0.37} & 0.67 & \underline{0.54} & 0.36 & 0.67 & 0.52 & \underline{0.32} & 0.66 & 0.49 \\
         & \ctm{} & 0.31 & 0.61 & 0.48 & 0.32 & 0.63 & 0.49 & 0.30 & 0.65 & 0.47 & 0.28 & 0.66 & 0.46 \\
        \cline{1-14}
        \bottomrule
        \end{tabular}
        }
    \caption{Fixed hyperparameters}
    \end{subtable}
\caption{Average alignment metrics across 10 runs \emph{for unseen category labels}. 
Measured against gold labels at the lowest hierarchy level, $|V| = 15,000$.
Largest values in each column are \textbf{bolded}, which are
 significantly greater than unbolded values in a two-sided t-test ($p<0.05$);
 \underline{underlined} values have $p>0.05$.}
\label{tab:cluster-results-test}
\end{table*}

\subsection{Additional Results}\label{app:fixed-hparam}\label{app:test-alignment}

\paragraph{Fixed Hyperparameters.} In \cref{tab:stability-results-by-seed,tab:cluster-results-by-seed}, we report the equivalents of \cref{tab:stability-results,tab:cluster-results} when holding hyperparameters \emph{fixed}, rather than letting them vary.
We identify the hyperparameters for each model that achieve the highest average alignment metrics on across experimental contexts for one dataset, then use those hyperparameters to estimate models on the other dataset (hyperparameter values in \cref{sec:app:hparams}).
In this way, we follow a common paradigm in practical application of machine learning models: hyperparameters are determined based on an initial experimental context, then used in another.
Broadly, \mallet{} is more stable and better-aligned than its neural counterparts in this setup, although the difference is not as stark as when hyperparameters are allowed to vary.

\paragraph{Held-out data.} In \cref{tab:cluster-results-test}, we report the alignment metrics for unseen category labels.
To form the held-out data, we keep all high-level categories consistent between the training and held-out sets, but partition the low-level categories such that some are never seen during training (e.g., although documents from the high-level \texttt{architecture} category will be included in both splits, documents on \texttt{bridges} are only seen in training while those on \texttt{lighthouses} are held-out).
Here too, \mallet{} generally has the highest alignment metrics over experimental contexts.

\subsection{Details of \lda{} applications meta-analysis}\label{sec:app:lda-meta}

\begin{table}
\resizebox{\columnwidth}{!}{
\setlength\tabcolsep{4pt}
\begin{tabular}{@{}lllr@{}}
\toprule
\multicolumn{1}{c}{\textbf{Coding Question}}                                                                                                                                                                         & \multicolumn{1}{c}{\textbf{Answer}}                                     & \multicolumn{1}{c}{\textbf{Count}} &          \\ \midrule
\begin{tabular}[c]{@{}l@{}}Is \lda{} used for inductive discovery of categories \\ for human consumption?\end{tabular}                                                                                                 & Yes                                                                     & 47                                 & (94\%)   \\
\begin{tabular}[c]{@{}l@{}}Which \lda{} estimates were used for the above \\ discovery?\end{tabular}                                                                                                                            & \begin{tabular}[c]{@{}l@{}}$\bm{\text{B}}$ only\end{tabular}               & 33                                 & (70\%) \\
                                                                                                                                                                                                                     & \begin{tabular}[c]{@{}l@{}}Both $\bm{\text{B}}$ and $\bm{\Theta}$\end{tabular} & 11                                 & (23\%) \\
\addlinespace
\begin{tabular}[c]{@{}l@{}}Is \lda{} used to categorize or represent individual \\ units of text (using $\bm{\Theta}$ estimate)?\end{tabular}                                                                                & Yes                                                                     & 32                                 & (64\%)   \\
\addlinespace
\begin{tabular}[c]{@{}l@{}}Are human-readable code labels assigned to topics \\ (formal \emph{content analysis})?\end{tabular} & Yes                                                                     & 34                                 & (68\%)   \\
\addlinespace
\begin{tabular}[c]{@{}l@{}}Is the exact \lda{} implementation specified?\end{tabular}                                                                                                                 & Yes                                                                     & 27                                 & (54\%)   \\ 
\addlinespace
\begin{tabular}[c]{@{}l@{}}Field of study\end{tabular}                                                                                                                            & \begin{tabular}[c]{@{}l@{}}Medicine\end{tabular}               & 21                                 & (42\%) \\
                                                                                                                                                                                                                     & \begin{tabular}[c]{@{}l@{}}Sociology\end{tabular} & 9                                 & (18\%) \\
                                                                                                                                                                                                                     & \begin{tabular}[c]{@{}l@{}}Business\end{tabular} & 7                                 & (14\%) \\
                                                                                                                                                                                                                     & \begin{tabular}[c]{@{}l@{}}Political Science\end{tabular} & 7                                 & (14\%) \\ 
                                                                                                                                                                                                                     & \begin{tabular}[c]{@{}l@{}}Psychology\end{tabular} & 4                                 & (8\%) \\ 
                                                                                                                                                                                                                     & \begin{tabular}[c]{@{}l@{}}Economics\end{tabular} & 2                                 & (4\%) \\ 
                                                                                                                                                                                                                     & \begin{tabular}[c]{@{}l@{}}History\end{tabular} & 2                                 & (4\%) \\   
\bottomrule
\end{tabular}}
\caption{Meta-analysis of fifty topic modeling papers outside the field of computer science (denominator may change, as not all conditions are always applicable). Content analysis is the dominant use case for topic models. The reliability, validity, and reproducibility of \lda{} estimates is critical to this use-case.}
\label{tab:meta-analysis-lda}
\end{table}

\begin{table}
    \resizebox{\columnwidth}{!}{
    \begin{tabular}{lllrrrr}
    \toprule
     & Algo. & $d$ & $\lambda$ & $>$ Worst & Med. & Best \\
    \midrule
    Overall & $k$-med. & RBO & 1.00 & 100\% & 97\% & 52\% \\
    \addlinespace
    \mallet{} & $k$-med. & RBO & 1.00 & 100\% & 98\% & 55\% \\
    \scholar{} & $k$-med. & Jcd. & 0.25 & 100\% & 100\% & 66\% \\
    \scholarkd{} & Aggl. & RBO & 0.75 & 100\% & 99\% & 60\% \\
    \dvae{} & Aggl. & Jcd. & 0.25 & 100\% & 92\% & 29\% \\
    \ctm{} & Aggl. & Jcd. & 0.25 & 100\% & 100\% & 90\% \\
    \bottomrule
    \end{tabular}
    }\caption{Alignment metrics for ensembles of each model, and how often they improve over the worst, median, and best member of the ensemble across 80 evaluation settings.}\label{tab:ensemble-results}
\end{table}

Summary statistics of our meta-analysis of studies using \lda{} outside computer science are shown in Table~\ref{tab:meta-analysis-lda}. The major results were discussed in Section~\ref{subsec:tm_for_content_analysis}. We find that about half of the papers did not specify the exact \lda{} implementation they used in their study, which raises larger reproducibility concerns for scientific research. Note that one paper can be assigned multiple subject or fields of study by Semantic Scholar. All the papers used for the meta-analysis are shown in \cref{tab:lda-meta-full1,tab:lda-meta-full2}. 

\begin{table*}
\centering
\rotatebox{90}{
\resizebox{\textheight}{!}{
\begin{tabular}{@{}lllllll@{}}
\toprule
\multicolumn{1}{c}{\textbf{Source}} & \multicolumn{1}{c}{\textbf{\begin{tabular}[c]{@{}c@{}}Field of \\ Study\end{tabular}}} & \multicolumn{1}{c}{\textbf{\begin{tabular}[c]{@{}c@{}}Is LDA used for \\ inductive discovery of \\ categories for \\ human consumption?\end{tabular}}} & \multicolumn{1}{c}{\textbf{\begin{tabular}[c]{@{}c@{}}Which LDA estimates \\ were used for \\ the inductive discovery?\end{tabular}}} & \multicolumn{1}{c}{\textbf{\begin{tabular}[c]{@{}c@{}}Is LDA used to \\ categorize or represent \\ individual units of text \\ (using $\bm{\Theta}$ estimate)?\end{tabular}}} & \multicolumn{1}{c}{\textbf{\begin{tabular}[c]{@{}c@{}}Are human-readable code labels \\ assigned to topics \\ (formal content analysis)?\end{tabular}}} & \multicolumn{1}{c}{\textbf{\begin{tabular}[c]{@{}c@{}}Is the exact LDA \\ implementation specified?\end{tabular}}} \\ \midrule
\cite{Santos2019TheDT} & Political Science & N & N/A & Y & N & N \\
\cite{Markides2022ATC} & Medicine & Y & Both & Y & Y & Y \\
\cite{miller2019three} & Political Science & Y & Both & N & N & Y \\
\cite{Scarborough2021PlaceBA} & Business & Y & Topic-Word Only & Y & N & N \\
\cite{Jang2021TrackingCD} & Medicine & Y & Topic-Word Only & Y & Y & Y \\
\cite{Noble2021UsingTM} & Medicine & Y & Topic-Word Only & Y & Y & Y \\
\cite{Yamada2019DetectionAT} & History & Y & Both & Y & N & Y \\
\cite{Zhang2021MonitoringDT} & Medicine & Y & Topic-Word Only & Y & Y & N \\
\cite{lee2021building} & Medicine & Y & Topic-Word Only & N & Y & N \\
\cite{Lattimer2022ExploringWT} & Medicine & Y & Both & Y & Y & N \\
\cite{Madzk2022StateoftheartOA} & Medicine & Y & Topic-Word Only & Y & Y & Y \\
\cite{wehrheim2021turn} & History & Y & Topic-Word Only & Y & Y & Y \\
\cite{Marshall2021ContributionOO} & Psychology & Y & Topic-Word Only & N & Y & N \\
\cite{Hopkins2020FromMD} & Political Science & Y & Topic-Word Only & Y & Y & N \\
\cite{Ning2020PredictionCT} & Medicine & Y & Topic-Word Only & Y & Y & N \\
\cite{Aniss2021TheRO} & Psychology & Y & Topic-Word Only & N & Y & Y \\
\cite{Neresini2019ExploringSF} & Sociology & Y & Topic-Word Only & N & Y & N \\
\cite{coco2020monetary} & Economics & Y & Topic-Word Only & Y & N & N \\
\cite{Almquist2019UsingRE} & Sociology & Y & Both & Y & Y & Y \\
\cite{Lee2020PolicyAP} & Political Science & Y & Topic-Word Only & N & Y & Y \\
\cite{Li2021TheRO} & Business & Y & Both & Y & Y & Y \\
\cite{Hou2021PublicAA} & Medicine & Y & Topic-Word Only & Y & Y & N \\
\cite{He2022TheLO} & Medicine & Y & Topic-Word Only & Y & Y & Y \\
\cite{Wang2021IsYS} & Business & Y & Topic-Word Only & N & N & Y \\
\cite{dybowski2019ecb} & Economics & Y & Both & Y & N & Y \\
 \bottomrule
\end{tabular}}}
\caption{Part one of all papers and their assessment for the meta-analysis of LDA use (Section~\ref{subsec:tm_for_content_analysis}).}
\label{tab:lda-meta-full1}
\end{table*}

\begin{table*}
\centering
\rotatebox{90}{
\resizebox{\textheight}{!}{
\begin{tabular}{@{}lllllll@{}}
\toprule
\multicolumn{1}{c}{\textbf{Source}} & \multicolumn{1}{c}{\textbf{\begin{tabular}[c]{@{}c@{}}Field of \\ Study\end{tabular}}} & \multicolumn{1}{c}{\textbf{\begin{tabular}[c]{@{}c@{}}Is LDA used for \\ inductive discovery of \\ categories for \\ human consumption?\end{tabular}}} & \multicolumn{1}{c}{\textbf{\begin{tabular}[c]{@{}c@{}}Which LDA estimates \\ were used for \\ the inductive discovery?\end{tabular}}} & \multicolumn{1}{c}{\textbf{\begin{tabular}[c]{@{}c@{}}Is LDA used to \\ categorize or represent \\ individual units of text \\ (using $\bm{\Theta}$ estimate)?\end{tabular}}} & \multicolumn{1}{c}{\textbf{\begin{tabular}[c]{@{}c@{}}Are human-readable code labels \\ assigned to topics \\ (formal content analysis)?\end{tabular}}} & \multicolumn{1}{c}{\textbf{\begin{tabular}[c]{@{}c@{}}Is the exact LDA \\ implementation specified?\end{tabular}}} \\ \midrule
\cite{gupta-etal-2020-heart} & Psychology & Y & Topic-Word Only & N & Y & Y \\
\cite{amozegar2021tweeting} & Business & Y & Topic-Word Only & Y & Y & Y \\
\cite{Bell2020CommonID} & Business & Y & Topic-Word Only & Y & N & N \\
\cite{vinan2021analyzing} & Sociology & Y & Topic-Word Only & N & N & Y \\
\cite{Yang2021HowDC} & Sociology & Y & Both & N & Y & N \\
\cite{ChandraDas2020ANALYSISOC} & Political Science & Y & Both & Y & Y & Y \\
\cite{Virtanen2021UncoveringDT} & Medicine & Y & Topic-Word Only & Y & Y & Y \\
\cite{ye2020people} & Sociology & Y & Topic-Word Only & Y & Y & N \\
\cite{Jing2021UserEA} & Medicine & Y & Topic-Word Only & N & N & N \\
\cite{Peterson2020WhatID} & Medicine,Sociology & N & N/A & Y & N & Y \\
\cite{Ko2021TheCO} & Political Science & Y & Doc-Topic Only & Y & Y & N \\
\cite{Mitchell2020Mode2KP} & Sociology & Y & Topic-Word Only & N & N & N \\
\cite{fernandez2021sentiment} & Business & Y & Both & N & Y & N \\
\cite{Bateman2019MeatAB} & Political Science & Y & Doc-Topic Only & Y & Y & Y \\
\cite{Fung2021EnhancedIP} & Medicine & Y & Topic-Word Only & Y & Y & Y \\
\cite{Yuan2020UnderstandingTE} & Sociology & Y & Topic-Word Only & Y & Y & N \\
\cite{Eom2021MarketableVE} & Medicine & Y & Topic-Word Only & N & N & Y \\
\cite{johnson2022herpes} & Medicine & Y & Topic-Word Only & Y & Y & N \\
\cite{Squires2022ShouldIS} & Medicine & N & N/A & N & N & N \\
\cite{Hu2021RevealingPO} & Medicine & Y & Topic-Word Only & Y & Y & N \\
\cite{Ng2021RoleBasedFO} & Medicine & Y & Topic-Word Only & N & N & N \\
\cite{Lee2021SWOTAHPAO} & Business & Y & Topic-Word Only & N & Y & Y \\
\cite{Mamaysky2020NewsAM} & Medicine & Y & Topic-Word Only & Y & Y & Y \\
\cite{Shin2019MultipleChoiceID} & Medicine, Psychology & Y & Both & N & N & Y \\
\cite{Park2020TwentyYO} & Sociology & Y & Doc-Topic Only & Y & Y & Y \\
 \bottomrule
\end{tabular}}}
\caption{Part two of all papers and their assessment for the meta-analysis of LDA use (Section~\ref{subsec:tm_for_content_analysis}).}
\label{tab:lda-meta-full2}
\end{table*}

\subsection{Hyperparameters}\label{sec:app:hparams}

Hyperparameters are included in the supplementary materials as \texttt{<model name>.yml} files. The full range of hyperparameters can also be found in Table \ref{tab:hyperparams}.

\begin{table*}[ht!]
    \footnotesize
    \centering
    \begin{subtable}{\textwidth}
		\centering
		\begin{tabular}{c c c c}
			\toprule
			\multicolumn{4}{c}{Model: \textbf{\mallet{}} } \\
		
		$\alpha$ & $\beta$ & Optim. Interval & $\#\text{Steps}$	                            \\
			\midrule
		$\{0.01, 0.05, 0.1, 0.25, 1.0^{*\dagger}\}$ & $\{0.01, 0.05^{*}, 0.1^{\dagger}\}$ & $\{0, 10^{*\dagger}, 500\}$ & $\{2000\}$                 \\
			\bottomrule
		\end{tabular}
		\caption{Hyperparameter ranges for \mallet{}. $\alpha$ is the topic density parameter. $\beta$ is the word density parameter. Optim. Interval sets the number of iterations between Mallet's own internal hyperparameter updates. $\#\text{Steps}$ are training iterations.}
		\label{tab:mallet_hyperparam}
    \end{subtable}

    \begin{subtable}{\textwidth}
		\centering
		\begin{tabular}{c c c c}
			\toprule
			\multicolumn{4}{c}{Model: \textbf{\scholar{}}} \\
		$\alpha$ & $\eta$ & $\eta_{BN}$ & $\#\text{Steps}$	                        \\
			\midrule
		\shortstack{$\{0.001, 0.005, 0.01,  0.5^{\dagger}, 1.0^{*}\}$ }& \shortstack{$\{0.001^{*\dagger},  0.002\}$} & \shortstack{$\{0.25^{*}, 0.5^{\dagger},  0.75\}$} & \shortstack{$\{200^{\dagger}, 500^{*}\}$}                 \\
			\bottomrule
		\end{tabular}
		\caption{Hyperparameter ranges for \scholar{}. $\alpha$ is the Dirichlet prior. $\eta$ is the learning rate. $\eta_{BN}$ is the epoch when batch-norm annealing ends (i.e., $\eta \times \text{Steps}$). $\#\text{Steps}$ are training epochs.}
		\label{tab:scholar_hyperparms}
    \end{subtable}
    
    \begin{subtable}{\textwidth}
		\centering
		\begin{tabular}{c c c c c c c}
			\toprule
			\multicolumn{7}{c}{Model: \textbf{\scholarkd{}}} \\
		$\alpha$ & $\eta$ & $\eta_{BN}$ & clipping & $T$ & $\lambda$ & $\#\text{Steps}$	                        \\
			\midrule
		\shortstack{$\{0.001, 0.005,$\\ $ 0.01,  0.5^{\dagger}, 1.0^{*}\}$ }& \shortstack{$\{0.001^{*},$ \\ $0.002^{\dagger}\}$} & \shortstack{$\{0.25, 0.5^{*},  0.75^{\dagger}\}$} & \shortstack{$\{0.0^{*}, 1.0^{\dagger},  2.0\}$} & \shortstack{$\{1.0^{\dagger}, 2.0^{*}\}$} & \shortstack{$\{0.5^{*},$ \\ $0.75, 0.99^{\dagger}\}$} & \shortstack{$\{200^{\dagger}, 500^{*}\}$}                 \\
			\bottomrule
		\end{tabular}
		\caption{Hyperparameter ranges for \scholarkd{}. $\alpha$ is the Dirichlet prior. $\eta$ is the learning rate. $\eta_{BN}$ is the epoch when batch-norm annealing ends. $\lambda$ is weight on the teacher model logits, $T$ is the softmax temperature, and clipping controls how much of the logit distribution to clip. $\#\text{Steps}$ are training epochs.}
		\label{tab:schlr-kd_hyperparam}
    \end{subtable}

    \begin{subtable}{\textwidth}
		\centering
		\resizebox{\textwidth}{!}{
		\begin{tabular}{c c c c c c}
			\toprule
			\multicolumn{6}{c}{Model: \textbf{\dvae{}}} \\
		$\alpha$ & $\eta$ & $\beta_{reg.}$ & $\gamma_{BN}$ & $\gamma_{KL}$ & $\#\text{Steps}$	                        \\
			\midrule
		\shortstack{$\{0.001^{\dagger}, 0.01^{*},  0.1\}$ }& \shortstack{$\{0.001^{\dagger},  0.01^{*}\}$} & \shortstack{$\{0.0, 0.01,  0.1^{*}, 1.0^{\dagger}\}$} & \shortstack{$\{0, 1,  100^{*}, 200^{\dagger}\}$} & \shortstack{$\{100^{*\dagger},  200\}$} & \shortstack{$\{500\}$}                 \\
			\bottomrule
		\end{tabular}
		}
		\caption{Hyperparameter ranges for \dvae{}. $\alpha$ is the Dirichlet prior. $\eta$ is the learning rate. $\beta_{reg.}$ is the $L_1$-regularization of the topic-word distribution. $\gamma_{BN}$ and $\gamma_{KL}$ are the number of epochs to anneal the batch normalization constant and KL divergence term in the loss, respectively. $\#\text{Steps}$ are training epochs.}
		\label{tab:dvae_hyperparam}
    \end{subtable}

    \begin{subtable}{\textwidth}
		\centering
		\begin{tabular}{c c c c}
			\toprule
			\multicolumn{4}{c}{Model: \textbf{\ctm{}}} \\
		
		$e(\cdot)$ & Learn Priors? & $\gamma_{\eta}$ & $\#\text{Steps}$	                            \\
			\midrule
		\shortstack{\{ \texttt{paraphrase-distilroberta-base-v2}, \\ \texttt{multi-qa-mpnet-base-dot-v1}$^{*\dagger}$, \\ \texttt{all-mpnet-base-v2} \}} & $\{\text{False}^{\dagger}, \text{True}^{*} \}$ & $\{0.001^{*}, 0.002\}$ & $\{100, 200^{*\dagger}\}$                 \\
			\bottomrule
		\end{tabular}
		\caption{Hyperparameter ranges for \ctm{}. $e(\cdot)$ is the Sentence Transformers document-embedding model\cite{reimers-gurevych-2019-sentence}. $\eta$ is the learning rate. $W_{decay}$ is the $L_2$ regularization constant. $\gamma_{\eta}$ is an indicator of whether learning rate is annealed. $\#\text{Steps}$ are training epochs.}
		\label{tab:ctm_hyperparam}
    \end{subtable}
    
\caption{Hyperparameter settings for \abr{MALLET}, \abr{D-VAE}, \abr{CTM}, \abr{SCHLR+KD} and \abr{SCHOLAR}. $*$: Best setting for \texttt{Bills}, $\dagger$: best setting for \texttt{Wiki}; based on the best average alignment metrics across experimental contexts.}\label{tab:hyperparams}
\end{table*}

\subsection{Additional alignment results}\label{sec:app:alignment}

\begin{figure*}
	\centering
	\begin{subfigure}[b]{0.95\textwidth}
	\includegraphics[width=1\linewidth]{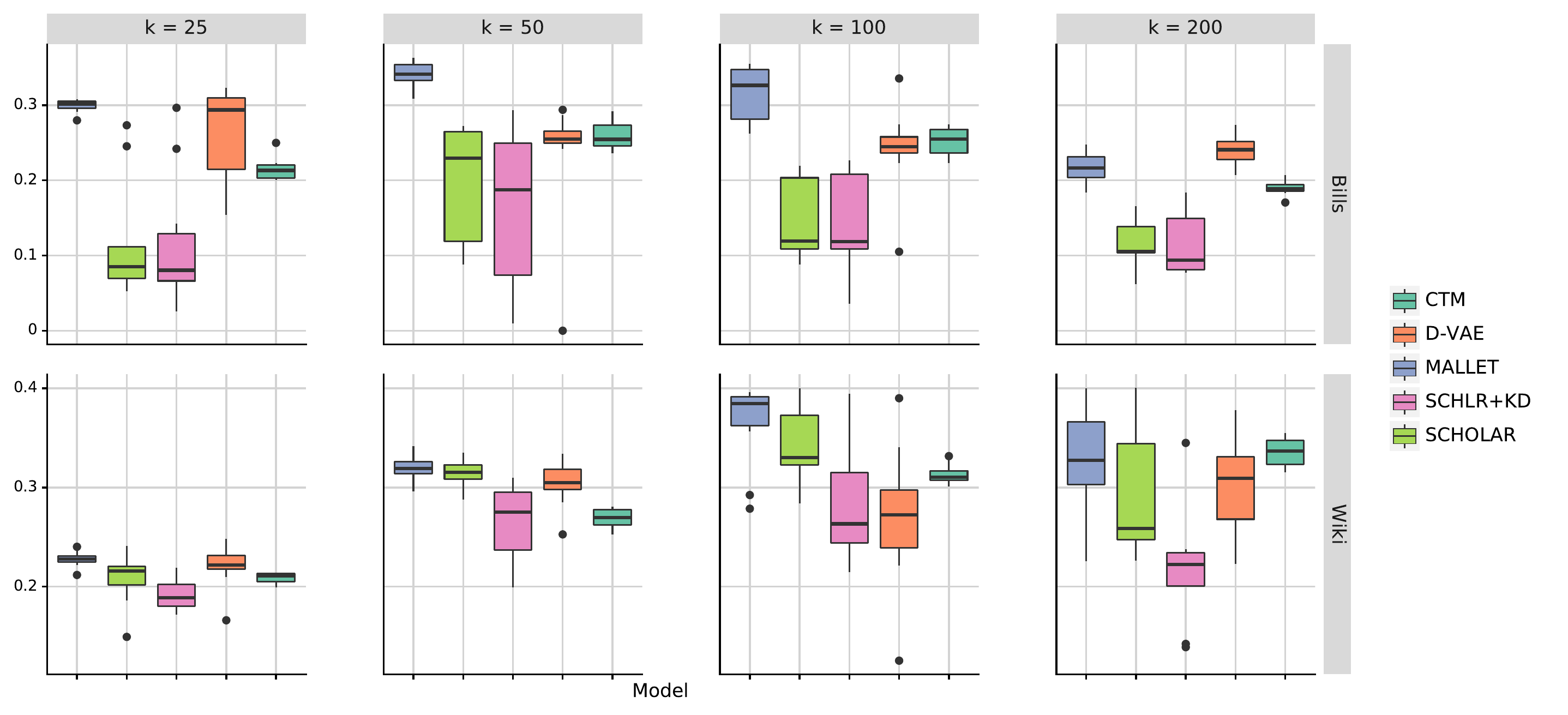}
	\caption{Alignment metric $=$ ARI}
	\end{subfigure}
	
	\begin{subfigure}[b]{0.95\textwidth}
	\includegraphics[width=1\linewidth]{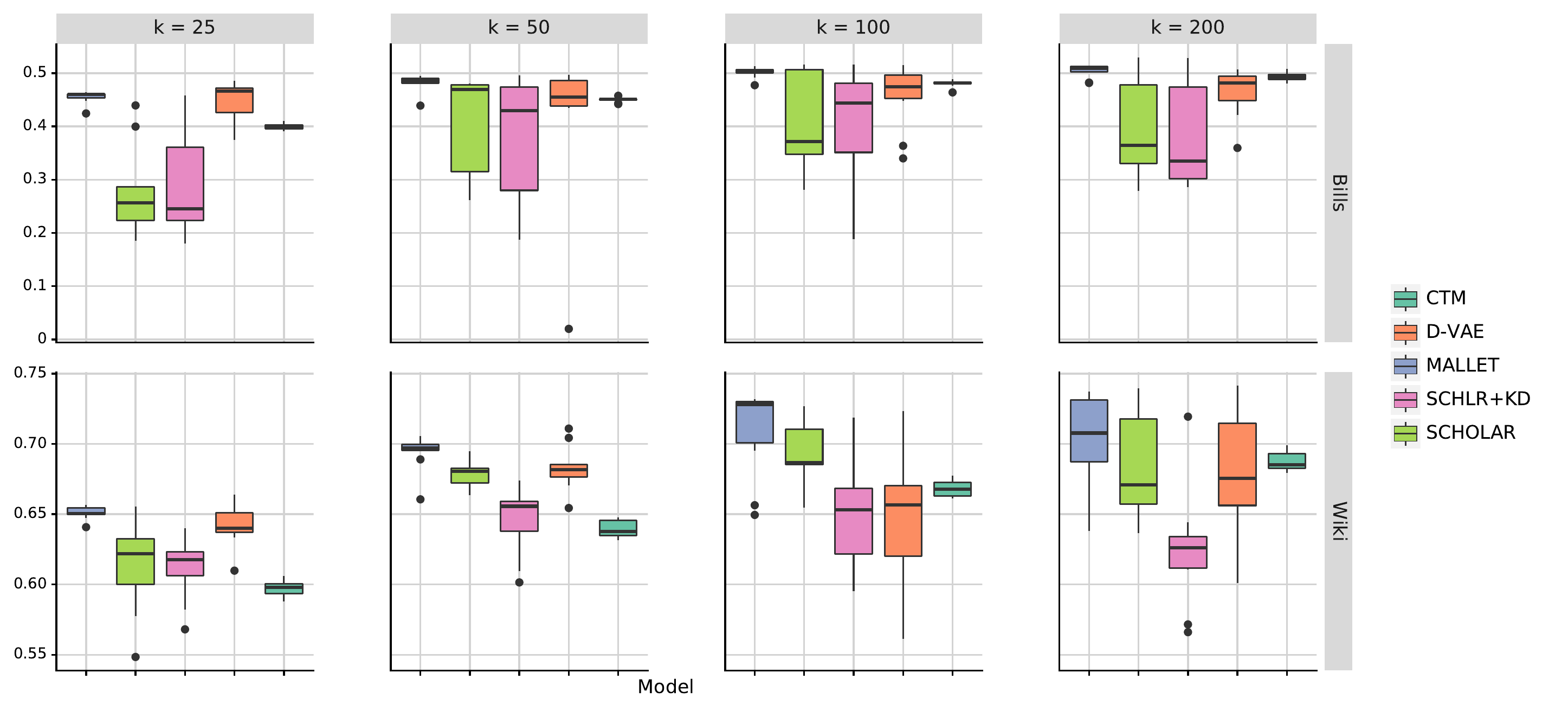}
	\caption{Alignment metric $=$ NMI}
	\end{subfigure}
	
	\begin{subfigure}[b]{0.95\textwidth}
	\includegraphics[width=1\linewidth]{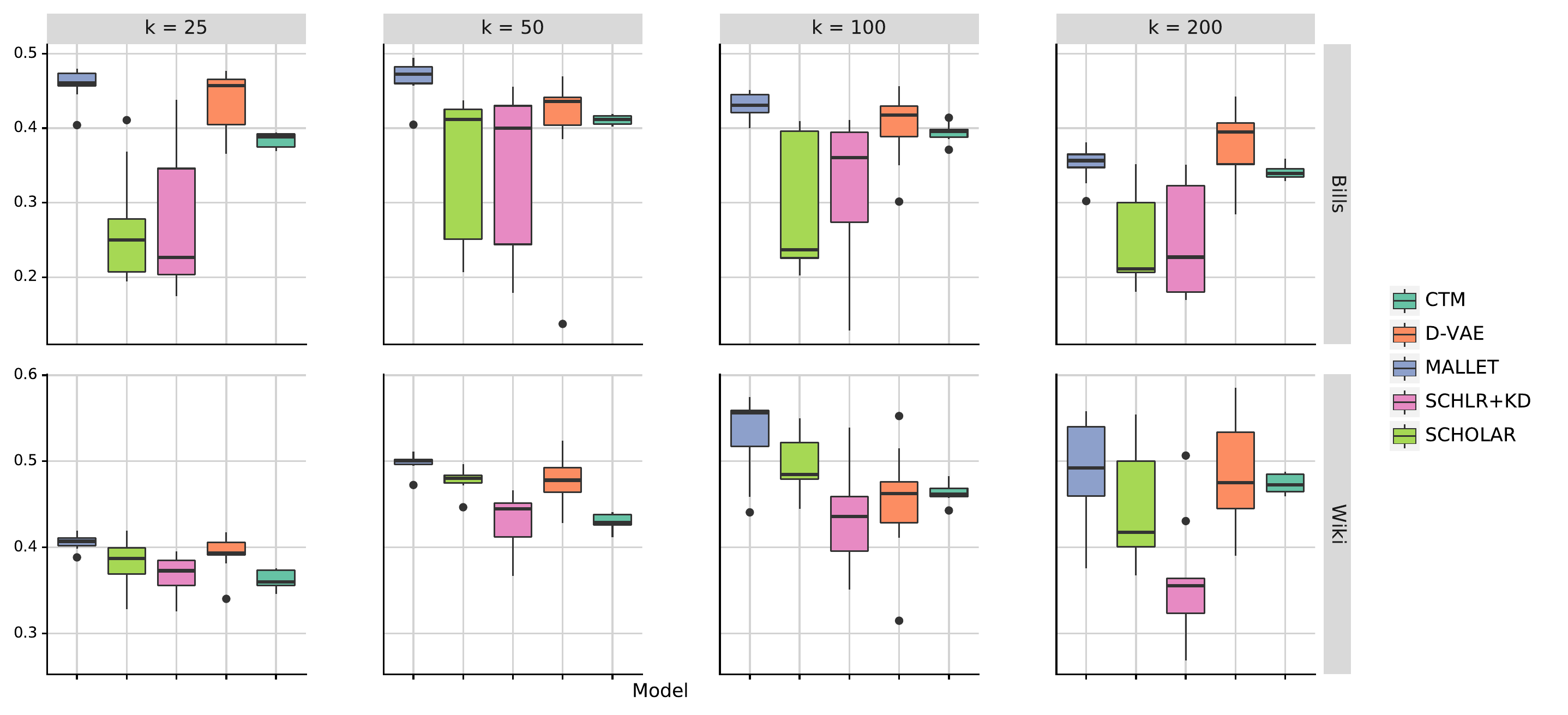}
	\caption{Alignment metric $=$ $P_1$}
	\end{subfigure}
	
\caption{
    Training set alignment metrics across five models,
        measured against gold labels at the lowest hierarchy level, $|V| = 15,000$.
}
\label{fig:cluster_qual_v15k_low}
\end{figure*}

\begin{figure*}
	\centering
	\begin{subfigure}[b]{0.95\textwidth}
	\includegraphics[width=1\linewidth]{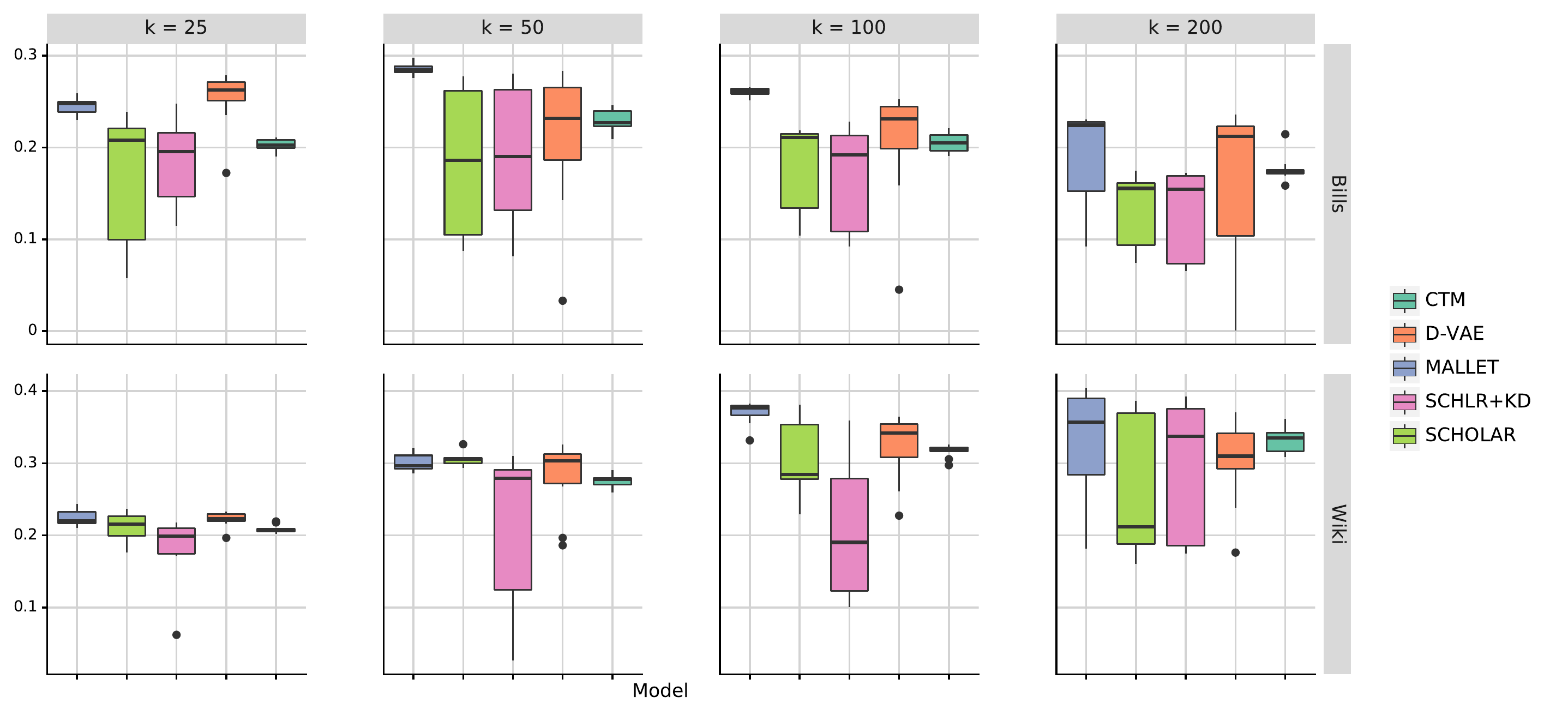}
	\caption{Alignment metric $=$ ARI}
	\end{subfigure}
	
	\begin{subfigure}[b]{0.95\textwidth}
	\includegraphics[width=1\linewidth]{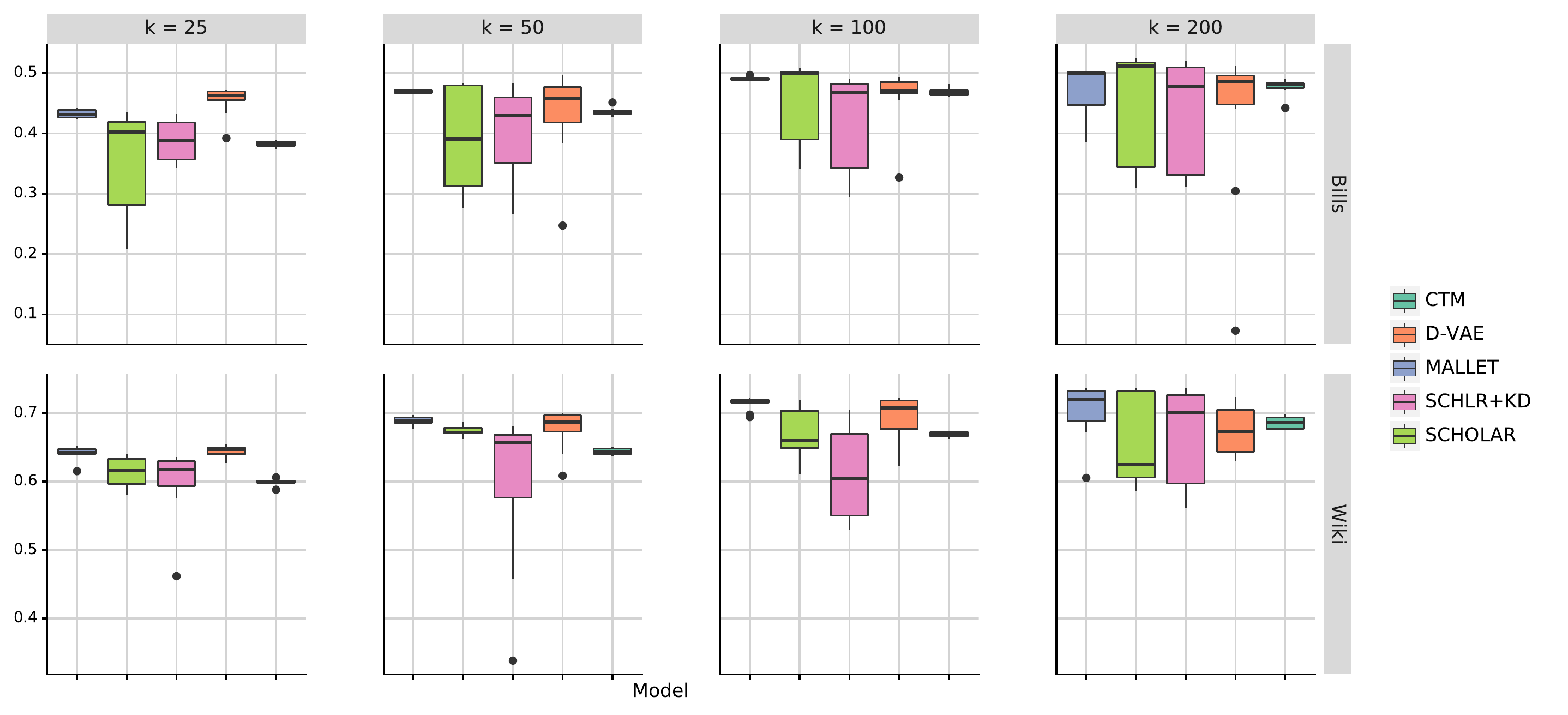}
	\caption{Alignment metric $=$ NMI}
	\end{subfigure}
	
	\begin{subfigure}[b]{0.95\textwidth}
	\includegraphics[width=1\linewidth]{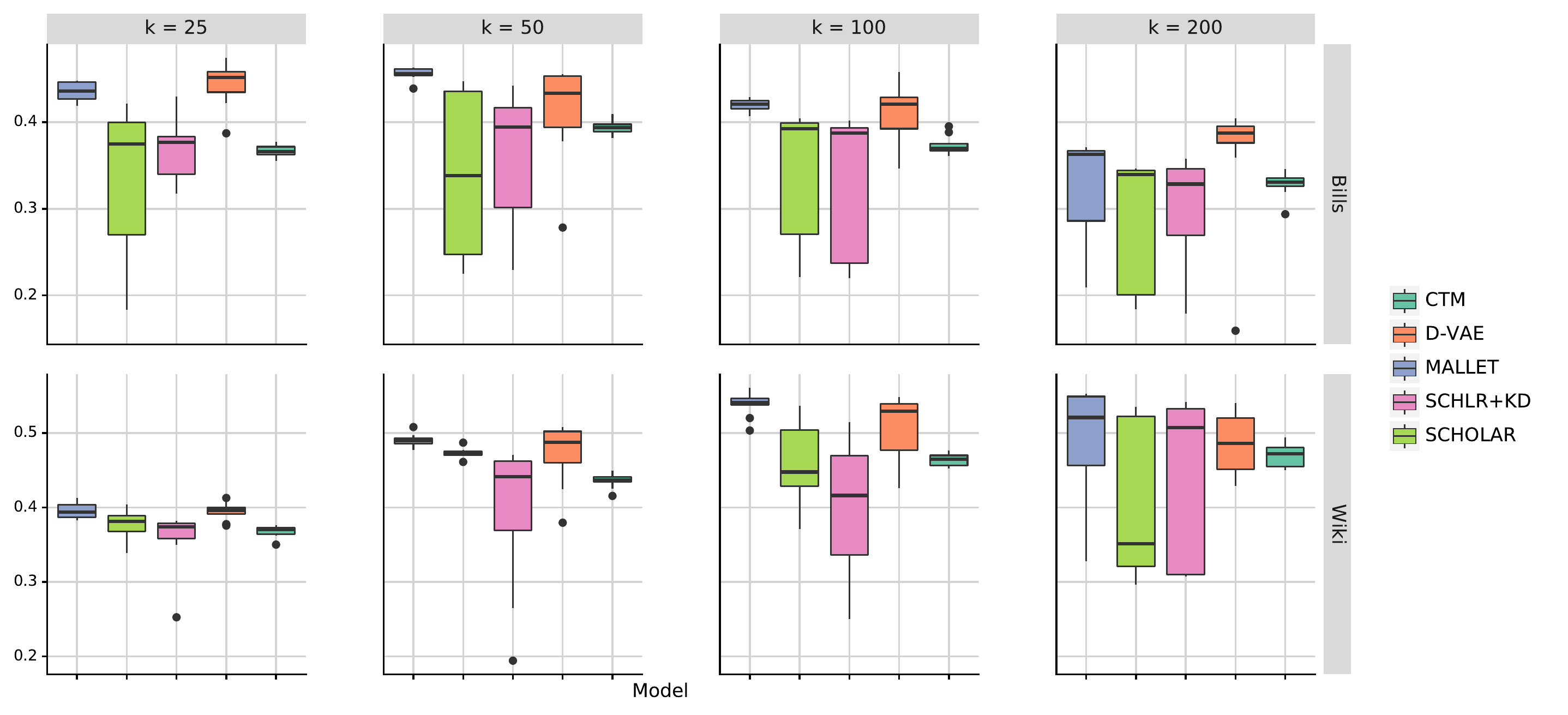}
	\caption{Alignment metric $=$ $P_1$}
	\end{subfigure}
	
\caption{
    Training set alignment metrics across five models,
        measured against gold labels at the lowest hierarchy level, $|V| = 5,000$.
}
\label{fig:cluster_qual_v5k_low}
\end{figure*}

\begin{figure*}
	\centering
	\begin{subfigure}[b]{0.95\textwidth}
	\includegraphics[width=1\linewidth]{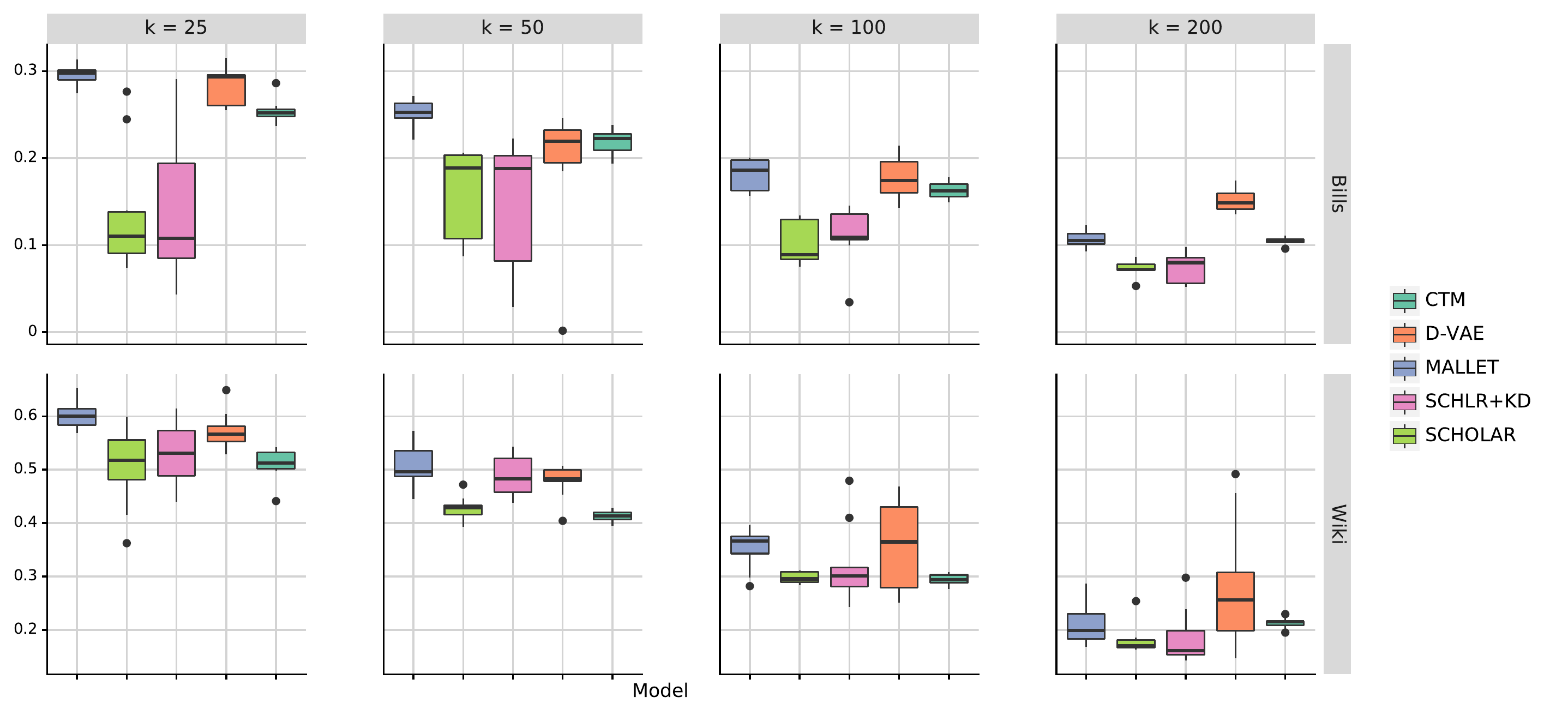}
	\caption{Alignment metric $=$ ARI}
	\end{subfigure}
	
	\begin{subfigure}[b]{0.95\textwidth}
	\includegraphics[width=1\linewidth]{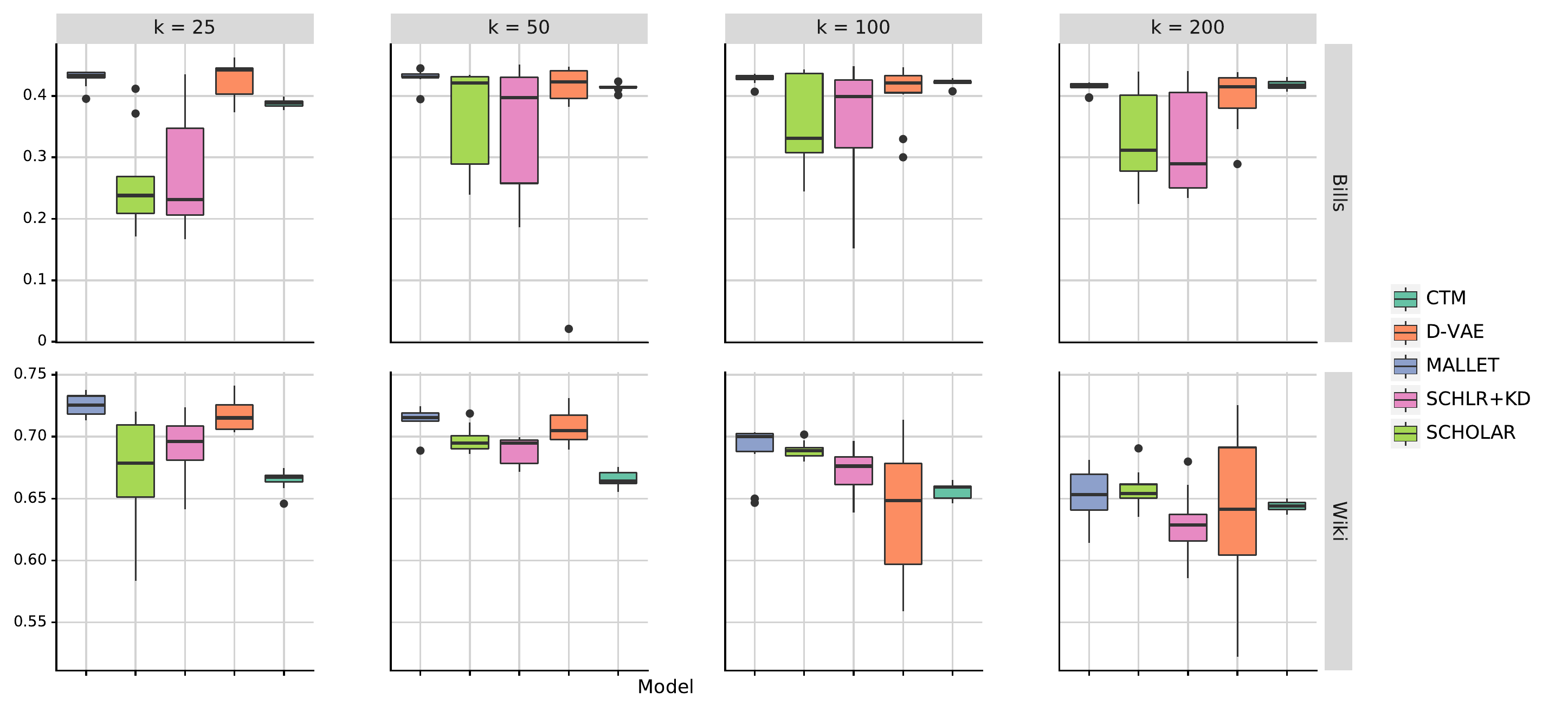}
	\caption{Alignment metric $=$ NMI}
	\end{subfigure}
	
	\begin{subfigure}[b]{0.95\textwidth}
	\includegraphics[width=1\linewidth]{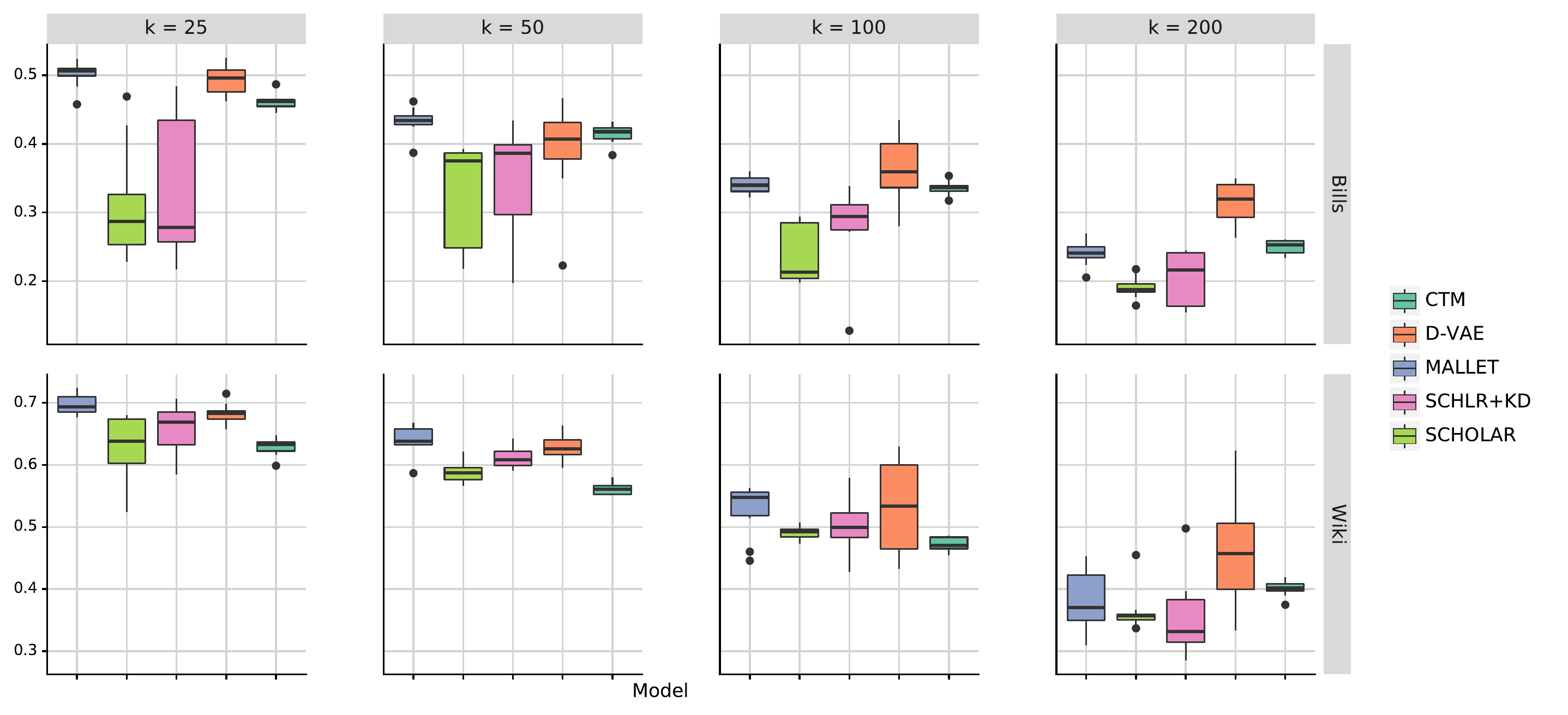}
	\caption{Alignment metric $=$ $P_1$}
	\end{subfigure}
	
\caption{
    Training set alignment metrics across five models,
        measured against gold labels at the highest hierarchy level, $|V| = 15,000$.
}
\label{fig:cluster_qual_v15k_high}
\end{figure*}

\begin{figure*}
	\centering
	\begin{subfigure}[b]{0.95\textwidth}
	\includegraphics[width=1\linewidth]{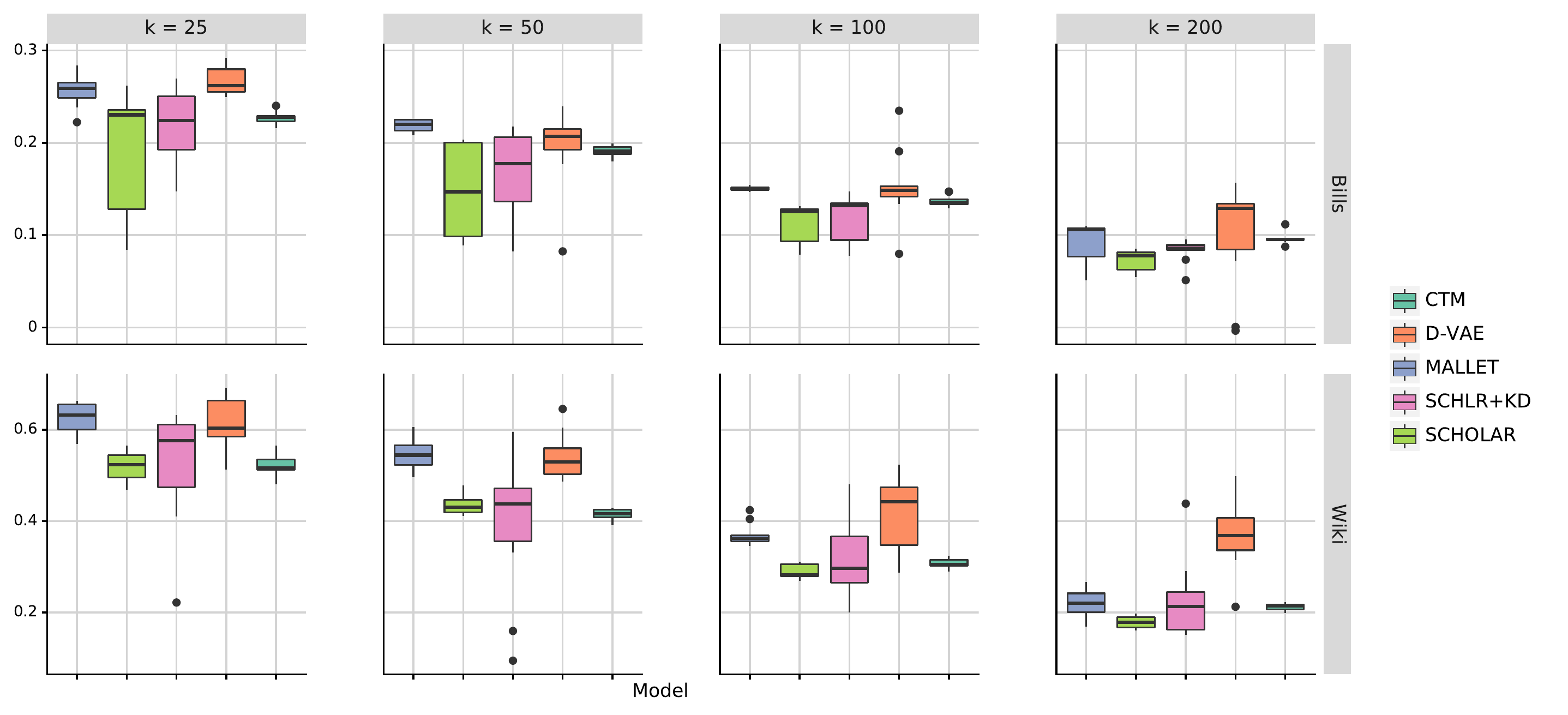}
	\caption{Alignment metric $=$ ARI}
	\end{subfigure}
	
	\begin{subfigure}[b]{0.95\textwidth}
	\includegraphics[width=1\linewidth]{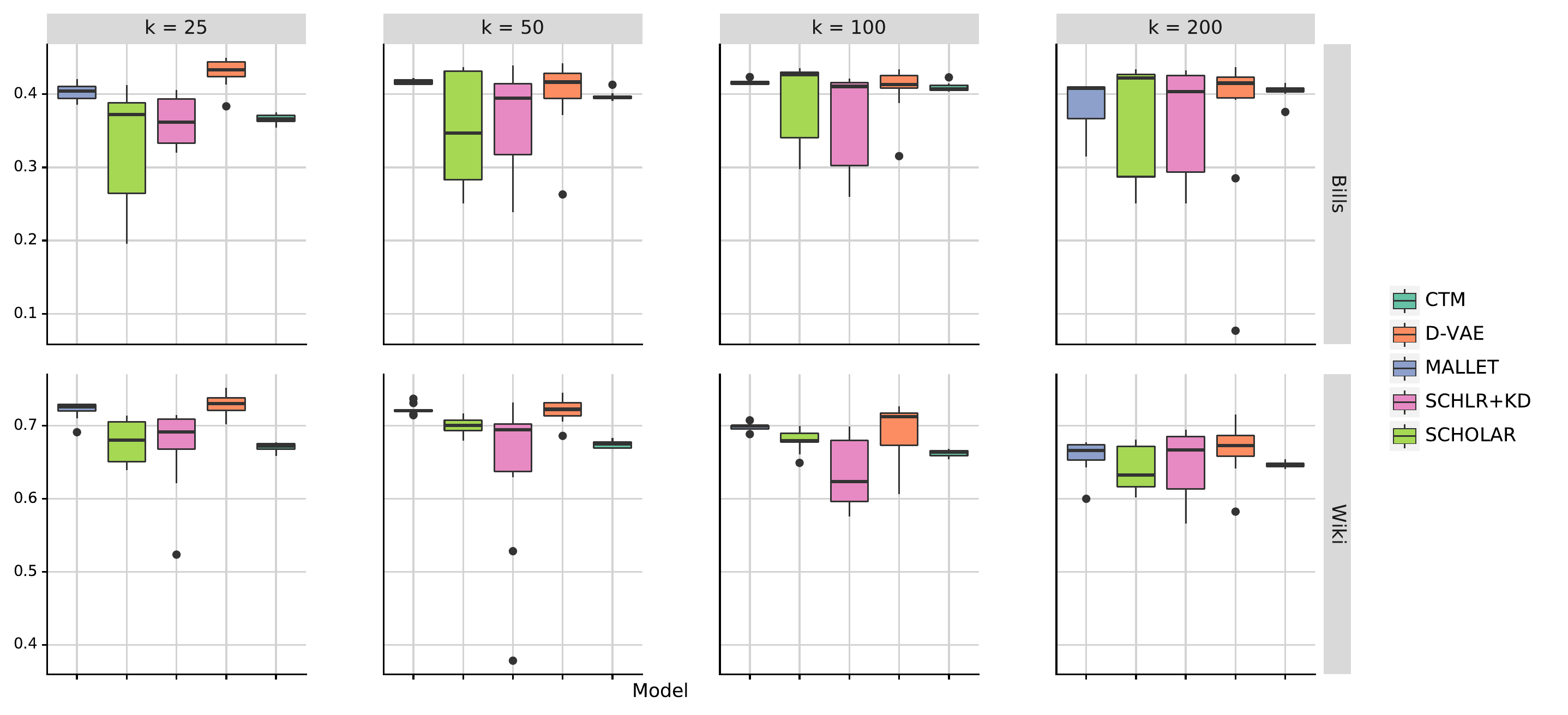}
	\caption{Alignment metric $=$ NMI}
	\end{subfigure}
	
	\begin{subfigure}[b]{0.95\textwidth}
	\includegraphics[width=1\linewidth]{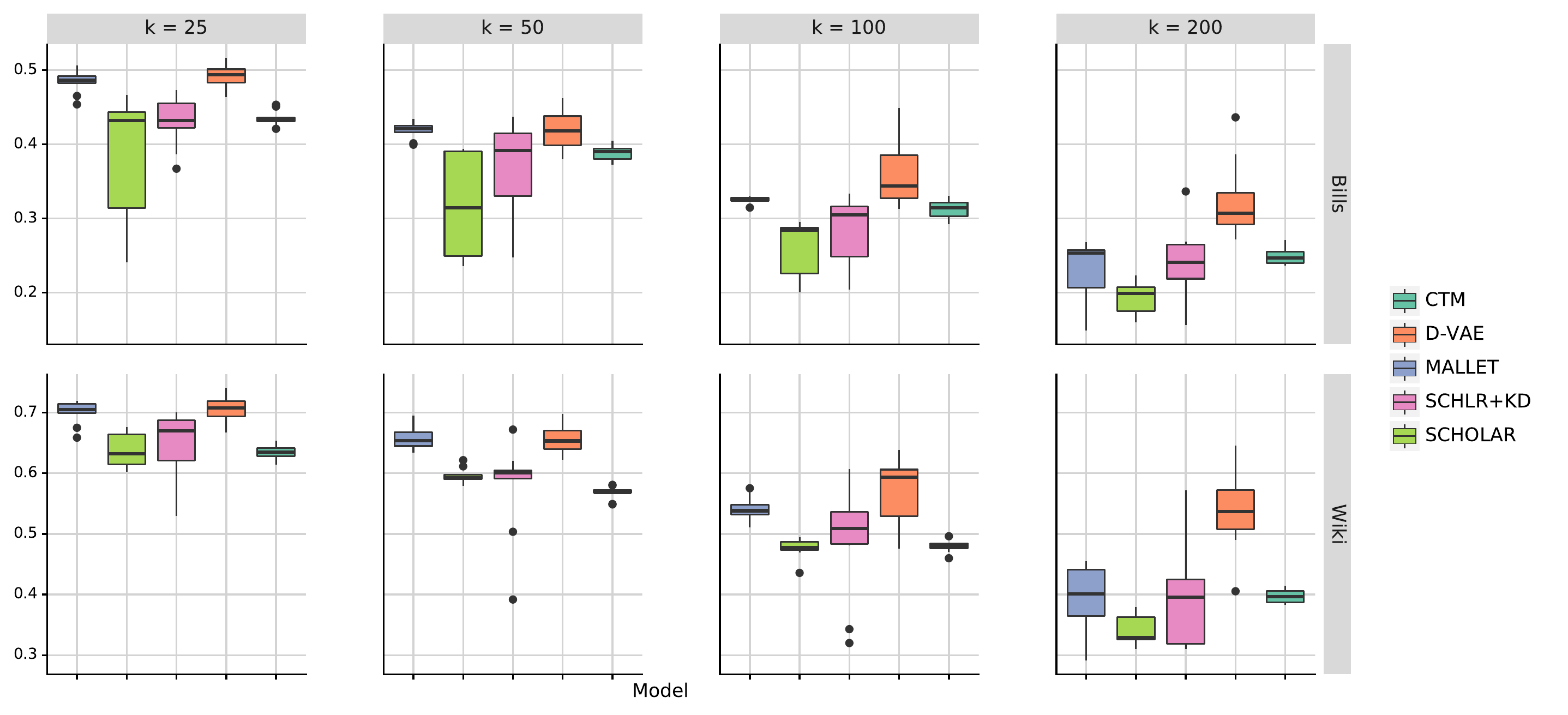}
	\caption{Alignment metric $=$ $P_1$}
	\end{subfigure}
	
\caption{
    Training set alignment metrics across five models,
        measured against gold labels at the highest hierarchy level, $|V| = 5,000$.
}
\label{fig:cluster_qual_v5k_high}
\end{figure*}

Results summarized in Table~\ref{tab:cluster-results} are shown in figure~\ref{fig:cluster_qual_v15k_low}. Results for rest of the settings for vocabulary and label hierarchy level are shown in \cref{fig:cluster_qual_v5k_low,fig:cluster_qual_v15k_high,fig:cluster_qual_v5k_high}.

\subsection{Additional ensembling results}\label{sec:app:ensemble}

In Table~\ref{tab:ensemble-results}, we list the best-performing ensemble per model type, alongside a method that fares well across all models.
For two out of the five models, the ensemble outperforms the median member in all 40 settings. Most ensembles improve upon the best member at least half the time.
We also identify a  set of hyperparameters (distance metric, $\lambda$, and clustering algorithm) that can ensemble the results of any of our models (\emph{Overall} row in Table~\ref{tab:ensemble-results}).

\subsection{Compute infrastructure}\label{sec:app:compute}
We used AWS ParallelCluster to provide a cloud-computing computing cluster. Neural topic models ran on NVIDIA T4 GPUs using \texttt{g4dn.xlarge} instances with 16 GiB memory and 4 CPUs.\footnote{\url{https://aws.amazon.com/hpc/parallelcluster/}}
\mallet{} ran on CPU only, with \texttt{m5d.2xlarge} instances (with 32 GiB memory, 8 CPUs).\footnote{See \url{https://aws.amazon.com/ec2/instance-types/} for further details.}

\end{document}